\pdfoutput=1

\documentclass[11pt]{article}

\usepackage[final]{acl}
\usepackage{amsmath} 
\usepackage{times}
\usepackage{stmaryrd}
\usepackage{latexsym}

\usepackage[T1]{fontenc}

\usepackage[utf8]{inputenc}

\usepackage{microtype}

\usepackage{inconsolata}
\usepackage{graphicx}
\usepackage{multirow}
\usepackage{booktabs}
\usepackage{algorithm}
\usepackage{algpseudocode}
\usepackage[switch]{lineno}
\usepackage{tcolorbox}
\usepackage{wrapfig}
\usepackage{subfig}
\usepackage{times}
\usepackage{amssymb}
\title{SRAP-Agent: Simulating and Optimizing Scarce Resource Allocation Policy with LLM-based Agent}

\author{
Jiarui Ji$^1$,Yang Li$^{1}$, Hongtao Liu$^1$, Zhicheng Du$^1$, \\
{\bf Zhewei Wei}$^{1,2}$, {\bf Weiran Shen}$^{1,2}$, {\bf Qi Qi}$^{1,2}$, {\bf Yankai Lin}$^{1,2}$\thanks{\quad Corresponding Author.}\\
$^1$Gaoling School of Artificial Intelligence,  Renmin University of China, Beijing, China\\
$^2$Beijing Key Laboratory of Big Data Management and Analysis Methods, Beijing, China
}

\begin{document}
\maketitle
\begin{abstract}
    Public scarce resource allocation plays a crucial role in economics as it directly influences the efficiency and equity in society. Traditional studies including theoretical model-based, empirical study-based and simulation-based methods encounter limitations due to the idealized assumption of complete information and individual rationality, as well as constraints posed by limited available data. In this work, we propose an innovative framework, SRAP-Agent (Simulating and Optimizing Scarce Resource Allocation Policy with LLM-based Agent), which integrates Large Language Models (LLMs) into economic simulations, aiming to bridge the gap between theoretical models and real-world dynamics. Using public housing allocation scenarios as a case study, we conduct extensive policy simulation experiments to verify the feasibility and effectiveness of the SRAP-Agent and employ the Policy Optimization Algorithm with certain optimization objectives. The source code can be found in \url{https://github.com/jijiarui-cather/SRAPAgent_Framework}.
    
\end{abstract}

\section{Introduction}

\begin{figure*}[t]
    \centering
    \includegraphics[width=\linewidth]{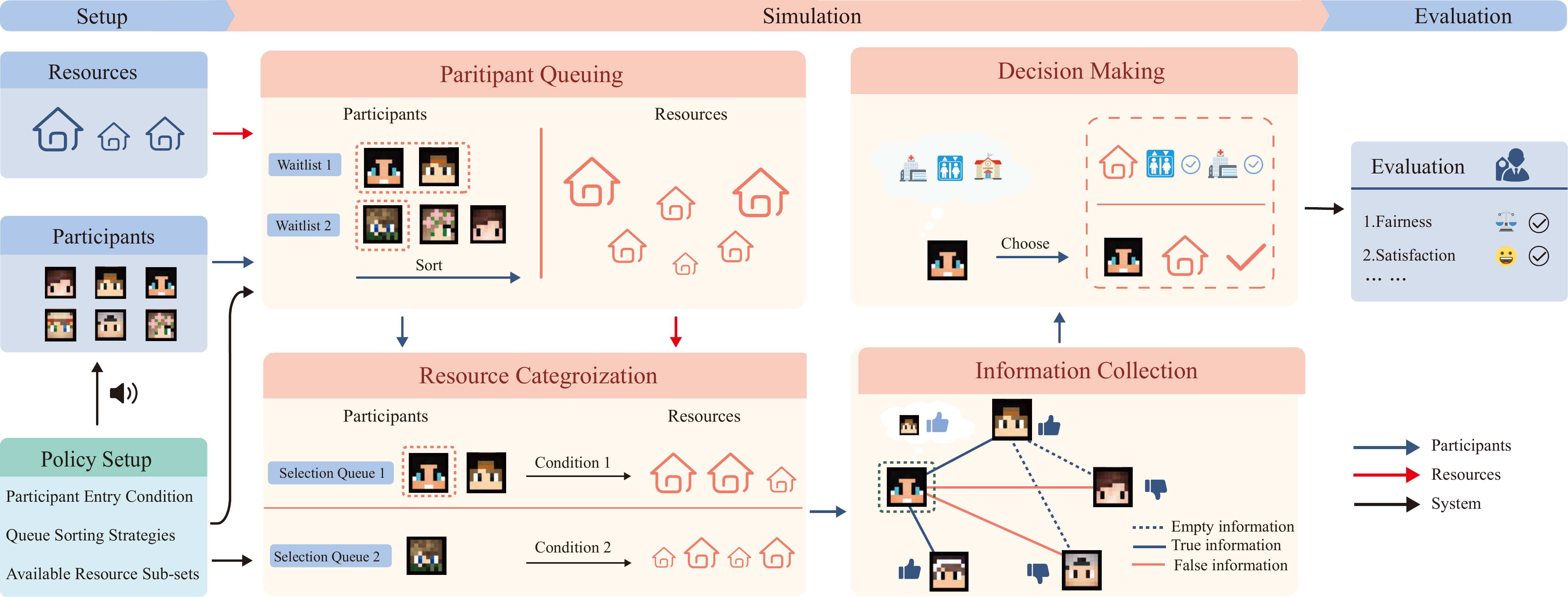}
    \caption{An illustration of the SRAP-Agent framework. The horn symbolizes the broadcasting process of policy information to participants.}
    \label{fig:eco_agent_framework}
\end{figure*}

Economics~\citep{mankiw2011} delves into how to limit scarce resources to meet unlimited needs. A critical aspect of this field is the allocation of public scarce goods~\citep{grove1977Optimal,groves1974incentive, BROCK1980}, focusing on utilizing limited resources to improve economic efficiency and social welfare ~\citep{wolfgang1986survey}. 
In the field of research on the allocation of public scarce resources, existing work can be primarily categorized into three main approaches: (1) theoretical model-based methods~\citep{hylland1979efficient,su2004patient,su2005patient,qi_waitlist}, which utilizes economic theories to develop models that can predict how resources should be allocated efficiently; (2) empirical study-based methods~\citep{BANERJEE2007287, Patricia1992, Josef2000}, which analyze real-world data to uncover patterns, correlations, and the effects of various policies; (3) computational simulation-based methods \citep{Artificial_adaptive_agents, Stephan2022aieco}, which emulate economic environments to test economic hypotheses within simulated settings. However, empirical study-based methods often face challenges of data scarcity, establishing causality, and isolating variables in complex social systems. Meanwhile, theoretical model-based methods and computational simulation-based methods often rely on simplified assumptions, overlooking the complex interplay of rational and social behaviors in human decision-making.

Fortunately, the emergence of LLMs such as ChatGPT and GPT-4 \citep{anil2023palm,touvron2023llama,brown2020language,openai2023gpt4}, has introduced a new potential in economic simulations. Acclaimed for their ability to mimic human-like behaviors~\citep{li2023camel, park2022social, park2023generative}, LLMs demonstrate the potential to encapsulate the logic and patterns inherent in human cognition by pre-training on extensive web data~\citep{openai2023gpt4}. This breakthrough facilitates the incorporation of social consensus mechanisms into economic simulations, offering a comprehensive framework for evaluating the impact of resource allocation policies on both economic efficiency and social welfare.

In this work, we develop an LLM-based scarce resource allocation policy simulation agent, SRAP-Agent. It abstracts the resource allocation queues for policy execution, with LLM-based agents carefully designed for the simulation of participants' behaviors. Furthermore, we propose the genetic algorithm-based policy optimization algorithm (POA) to find optimal policies towards pre-defined targets. Through experiments in the context of public housing allocation, we validate the effectiveness of SRAP-Agent, which reveals several key insights: (1) The LLM-based agent can effectively simulate the emotion factor and strategic behaviors of humans. Compared to the GPT-4-driven agent, the simulated decision-making behavior of the GPT-3.5-driven agent is more similar to humans' behavior. (2) In scarce resource allocation, pivotal factors include the entry conditions of participants, the queuing method, and the categorization of resources. (3) POA can optimize policies towards a specific policy evaluation metric, and improve approximately 20\% on this metric.

\section{Related Work}
\textbf{Public Scarce Resources Allocation.}
The allocation of public scarce resources represents a crucial area of inquiry in the development of economic policy. A multitude of researchers \cite{hylland1979efficient, shapley1974cores, sonmez1999strategy, abdulkadirouglu1999house} conduct in-depth investigations into static matching models and propose Pareto efficient, individually rational, and strategy-proof simple mechanisms for various scenarios. Later, given the predominantly dynamic nature of practical scarce resource allocation, several studies \cite{sonmez2011matching, su2004patient, su2005patient, bloch2017dynamic} begin to integrate dynamic properties into their models. 
The waitlist mechanism is widely adopted for allocating scarce resources. For instance, \citet{qi_waitlist} explore the waitlist mechanism in public housing allocation, \citet{agarwal2021equilibrium} design a waitlist for kidney organ allocation, which increased donor supply by 18.2\% to enhance patient welfare; \citet{lewis2000ending} develop a mechanism for managing waitlist for elective surgeries, improving the fairness of surgery opportunity distribution among patients. 
We focus our research on the k-deferrals waitlist mechanism in \cite{qi_waitlist}.


\textbf{Economic Policy Simulation.}
The predominant methodologies for simulating the implementation of economic policies include rule-driven and reinforcement learning-based simulations. Rule-driven approaches~\cite{Artificial_adaptive_agents, Bonabeau2002, Farmer2009TheEN}, involve modeling utility functions as agents, formulating various environmental rules, and observing the resultant global changes. Reinforcement learning-based simulations~\cite{Laurent2011NotMarkovian, Claus1998DRL, Stephan2022aieco, bansal2018emergent, Jaderberg_2019}, involve modeling agents as either Markov Decision Processes (MDP) or Partially Observable Markov Decision Processes (POMDP), employing reinforcement learning methods to maximize the utility of each agent. However, both methods are heavily dependent on the assumption of individual rationality, often leading to predictions that diverge from real-world outcomes. While exploring the same problem, these methods presuppose that agents act rationally and calculate behavior by first optimizing equilibrium outcomes.

Our research introduces a novel approach by integrating social consensus derived from LLMs with the conventional concept of individual rationality to generate choices that reflect individual preferences. We optimize policy based on simulation results, providing a unique research perspective incomparable to existing methods.

\textbf{LLM-based Agents. }
With the development of LLMs~\cite{openai2023gpt4,anil2023palm,touvron2023llama}, there has been an emerging focus on the exploration of LLMs-based agent architectures and prompt designs ~\cite{kojima2023large, shinn2023reflexion, wang2023describe}, aimed at augmenting the capabilities of LLM-based agents to perform more complex tasks. Alongside these developments, researchers are devoted to crafting realistic societal simulation environments~\cite{li2023camel, park2022social, park2023generative} that incorporate multiple agents, thereby providing a more dynamic and interconnected framework for agent interaction and behavior analysis. Following this, ~\citep{li2023large} makes a primary exploration of the effectiveness of LLMs-based agent simulation in macroeconomic policy. However, it remains unconfirmed whether the decision-making capabilities of LLMs align with those of actual humans in economic activities. Our investigation draws inspiration from the experimental findings of Dillion\cite{DILLION2023597}, aiming to integrate social consensus with individual rationality within the economic policy simulation process. 

\section{SRAP-Agent}

In this section, we present SRAP-Agent, a novel large language model (LLM) agent-based simulation framework designed for examining the impact of economic policies on public scarce resource allocation. SRAP-Agent aims to consider that human decision-making in economic contexts is not purely rational, but is influenced by a mix of rationality and emotion, leading to unpredictable outcomes.

The simulation of SRAP-Agent is divided into three phases as shown in Fig. \ref{fig:eco_agent_framework}: (1) {Setup phase}, involves the initialization of critical variables including the profiles of participants $P$, the information of available public scarce resources $R$, and the allocation policy $\pi$. (2) {Simulation phase}, dynamically simulates the allocation of resources $R$ among participants $P$, adhering to the constraints and rules defined by the allocation policy $\pi$. (3) {Evaluation phase}, focuses on assessing the outcomes of the allocation process, employing various metrics such as fairness and participant satisfaction.

Next, we outline the formulation of the allocation policy $\pi$ and the structure of participants $P$.  



\subsection{Allocation Policy}

Public scarce resource allocation represents a significant challenge in economics, particularly in the context of public resources where the objectives of efficiency and equity frequently converge. In this work, we consider the queuing, pricing, and grouping mechanisms in SRAP-Agent. Formally, In the SRAP-Agent framework, we formalize a structure consisting of $m$ distinct queues: $q_1, q_2, \ldots q_m$. Each queue is uniquely characterized by: 

(1) \textbf{Participant entry conditions}: to define the conditions under which individuals are eligible to enter a specific queue. Generally, the entry condition $E_{queue}(i)$ for the queue $q_i$ typically encompasses various socioeconomic factors such as individual budget, average income, and other relevant personal information. In the context of budget-based criteria, the entry conditions for $m$ queues are categorized into $m$ distinct budget ranges, organizing from high to low budget. This allows the simulation to reflect diverse economic backgrounds and their access to resources.

(2) \textbf{Queue sorting strategies}: to govern the prioritization of participants within each queue upon their entry, denoting as $S_{queue}$. We have incorporated two distinct sorting strategies: (a) first-in-first-out, of which participant order is set by arrival sequence. This method, prevalent in real scenarios, guarantees procedural fairness by chronologically allocating opportunities without regard to participants' characteristics. (b) priority for vulnerable groups, which is designed to address social equity by providing preferential access to vulnerable groups, positioning them at the forefront when entering the specific queue. Besides, SRAP-Agent also employs the widely-used waiting queue \cite{qi_waitlist, agarwal2021equilibrium} mechanism. Each queue encompasses two components: the waiting queue and the selection queue. 
 Participants initially enter the waiting queue. When a spot becomes available in the selection queue, the participant at the front of the waiting queue is transferred to the selection queue. The number of resources is denoted as $|R|$, then the capacity of the selection queue for accommodating participants is upper-bounded by $c \cdot |R|$. In the selection queue, each participant can queue to choose up to $k$ times; once their choices are exhausted, they return to the waiting queue. This mechanism diminishes the waiting time for participants by increasing the chances of participants staying in the selection queue.

(3) \textbf{Available resource sub-sets}: These sub-sets represent the specific resources that participants in a given queue can access and choose from, denoting as $R_{queue}(i)\subseteq R$ for the $q_i$. The composition of these resource sub-sets is dictated by certain predefined conditions, primarily considering factors such as the price and quality of resources. Typically, the predefined conditions of the resources in a queue are matched with their corresponding entry conditions. This ensures that resources are categorized and allocated to participant groups in a manner that is congruent with their economic capacity. 
Hence, the allocation policy is defined as the following equation:
\begin{equation}
    V(p_j) = \pi(p_j|E_{queue}, S_{queue}, R_{queue}, m),
    \label{equation:policy_setting}
\end{equation}
where $p_j$ represents the $j$-th participant in the list of participants $P = (p_1, p_2, \cdots, p_n)$, the function $V(p_j)$ is the available resource for the $j$-th participant under the policy. It enters the $q_i$ when it satisfies the condition $E_{queue}(i)$. Its order in the queue is determined by $S_{queue}$. When it is his turn to select, the participant $p_j$ can choose from the remaining resources in $R_{queue}(i)$, i.e., $V(p_j)$. In SRAP-Agent, the policymakers can set the resource policy according to their requirement, and the policy will be broadcast to all participants.

\subsection{LLM Agent-based Participant Simulation}

Recognizing the complexity of human decision-making, SRAP-Agent employs LLM-based agents to simulate participants' behaviors in the context of public scarce resource selection. This approach recognizes that human decision-making is not purely rational and often influenced by a range of factors. While it is impractical to simulate all aspects of human behavior, SRAP-Agent focuses on two most impactful behaviors: (1) Decision-making behavior. This aspect of the simulation addresses how participants evaluate and choose resources, factoring in elements like personal preferences, perceived value, and immediate needs. (2) Social behavior. Social dynamics also play a critical role in resource selection. This includes interactions with other participants, social norms, and external influences, which can affect how participants perceive and choose resources.

\paragraph{Agent Architecture} 
SRAP-Agent leverages LLM-based agents to instantiate individual participants in the simulation. Each participant $p_j$ is uniquely characterized by an initial profile including several key personal attributes such as economic status, income level, family background, and social network connections. To facilitate dynamic interaction and decision-making, each agent is equipped with a memory component $m_{j}$ describing the participant's activity history. This memory serves as a critical reference point, enabling the agent to make informed decisions based on past actions and interactions within the simulation environment. This design not only enhances the realism of the simulation but also allows for a more nuanced exploration of social dynamics and individual decision-making processes.




\paragraph{Social Behavior Simulation}

Human decision-making often involves the collection of information to make more informed choices. SRAP-Agent incorporates two primary social behaviors frequently employed by humans for information collection:

(1) \textbf{Broadcasting}: This mechanism is akin to a conventional web blog. Participants in SRAP-Agent utilize this platform for both disseminating and acquiring information, mirroring the way individuals interact and gather information in real-world social networks.


(2) \textbf{Private messaging}:  In addition to broadcasting, participants can send private messages to their friends. This feature is critical for more direct and personalized communication, such as seeking assistance or sharing specific information about high-quality resources. 
In human society, communication is deeply influenced by psychological factors that shape the information being shared, resulting in asymmetrical information within social networks.
To simulate the mental state of humans and the strategic interactions
among different participants, we adopt a mechanism of \textit{memory assessment}.
We design the memory $m_j$ of $p_{j}$ to comprise two parts: (1) \textbf{Trustworthy memory $m_j^T$} that mainly includes information from reliable sources, such as policymakers; (2) \textbf{Suspicious memory $m_j^S$} that stores information from the social behaviors. 

In the process of communication among participants, assume that $p_j$ is interacting with $p_k$, and the received information from $p_k$ will be stored in suspicious memory $m_j^S$ firstly. Then, $p_j$ measure the reliability of information from participant $p_k$ from two aspects: i) assessing the nature of their relationship with $p_k$, encompassing factors like closeness and historical interactions, in conjunction with a moral appraisal of $p_k$; ii) contrasting the received information against their trusted memory $m_j^T$. Following this evaluation, parts of the received information that are deemed reliable are integrated into the trustworthy memory $m_j^T$.





Through these mechanisms, SRAP-Agent effectively models the complex social dynamics of information gathering and sharing, allowing for a deeper understanding of human behavior in the context of public scarce resource allocation.

\paragraph{Decision-Making Behavior Simulation}

The simulation of the decision-making processes in SRAP-Agent can be formulated by:
\begin{equation}
    R^*_j = D(p_j, V(p_j)),
    \label{equation:makeing_decision}
\end{equation}
where the function $D(p_j, V(p_j))$ indicates the process that $p_j$ selects the desired resource from his visible resource pool. $R^*_j = \emptyset$ means participant $p_j$ quit the decision procedure when there's no desired resource.

\section{POA: Policy Optimization Algorithm}
\begin{algorithm}[t]
\caption{POA: Policy Optimization Algorithm}
\label{algorithm:POA}
\begin{algorithmic}[1]
\Require Historical policies $\Pi_{h}$, predictor $\widetilde{f}$, running iterations $M$ 
\Ensure Optimized policy $\pi^*$

\State Randomly generate policies $\Pi_{r}$.
\State Initialize policy pool $\Pi$ with $\Pi_{h}$ and $\Pi_{r}$.
\For{$i = 1$ to $M$}
    \State Calculate the fitness $\widetilde{f}(\pi)$ of $\pi$ in $\Pi$.
    \State Use the tournament selection \cite{miller1995genetic} to select policies $\Pi_{s}$ in $\Pi$ based on fitness.
    \State Combine pairs in $\Pi_{s}$ to produce $\Pi_{c}$. 
    \State Apply mutation to the $\Pi_{c}$ to obtain $\Pi_{m}$. 
    \State Replace $\Pi_{s}$ with $\Pi_{m}$ to update $\Pi$.
\EndFor
\State Obtain the optimized policy by $\widetilde{f}$: 
\begin{equation*}
    v_{\pi^*} = \underset{v_\pi}{\arg \max } \widetilde{f}\left(v_\pi\right), \pi \in \Pi
\end{equation*}
\State Decode $v_{\pi^*}$ to obtain the optimized policy $\pi^*$.
\end{algorithmic}
\end{algorithm}

We propose a policy optimization algorithm (POA) based on the genetic algorithm \cite{lambora2019genetic} to find more reasonable policies.
The set of policies proposed by the policymaker is $\pi \in \Pi$, and $f$ is the policy evaluation metric, then the optimal policy is
$\pi^{*}=\underset{\pi \in \Pi}{\arg \max } f(\pi).$
Generally, $f$ takes into account $L$ different kinds of policy evaluation metrics ($f_{j}(\pi), j \in [L]$), which can be categorized into societal satisfaction and societal fairness metrics. We pre-set the weights $w_j$ to alter the optimization objective of POA, the specific calculation formula of $f$ is as follows:
\begin{equation}
f(\pi) = \sum_{j=1}^n w_j \cdot f_j(\pi), \pi \in \Pi_{h}.
\label{equation:weight_poa}
\end{equation}
Due to the excessive number of policies that need to be searched, the time required to obtain policy evaluation results with SRAP-Agent may be excessively long. So we use the $v_{\pi}, f(\pi)$ dataset to train a \textit{predictor} $\widetilde{f}$ for the estimation of the policy evaluation result: $f(\pi) = \widetilde{f}(v_{\pi})$. POA then performs $M$ iterations to find more rational policy parameters. The specific steps of POA are listed in Algorithm.\ref{algorithm:POA}.

\section{Experiment}
We select the public housing allocation scenario as a typical case to validate the simulation effectiveness of SRAP-Agent.
 We conduct experiments in three steps:
 1. We conduct the Turing test to ensure that, SRAP-Agent can effectively simulate the policy execution process of various policies proposed by the policymaker.
 2. Based on the simulation results obtained from the SRAP-Agent, we analyze the impact of various policy parameters on the allocation of public scarce resources. 
 3. Subsequently, we employ the POA algorithm to find the optimal policy in pursuit of specific optimization objectives.




\subsection{Evaluation Metrics}


In our research, we adhere to the commonly used metrics for public scarce resources allocation \cite{metrics,survey_fairness}, employing two major categories of policy evaluation metrics: 

\textbf{Societal Satisfaction Metrics.}
We use three metrics to evaluate societal satisfaction: (1) Avg $r^{size}$: representing the average per-capita living area size of house resources for all participants; (2) Avg $WT$: Average waiting time for each participant; (3) \(SW\): the social welfare, quantifying the cumulative satisfaction of all participants. 

\textbf{Societal Fairness Metrics.}
In addition, we employ four metrics to evaluate societal fairness: (1) Var $r^{size}$: denoting the variance in per capita living area size among participants; (2) Rop: indicating the number of inverse order pairs in house allocation results; (3) co-Gini: the Gini coefficient \cite{survey_fairness} calculated on house allocation result. We use $F(V, NV)$ to reflect the \(SW\) gap between the vulnerable ($V$) and non-vulnerable group ($NV$) of participants.
The calculation equations are listed in Appendix~\ref{section:evaluation_metrics_equation}.

\subsection{Turing Test}
\begin{table}[!t]
\caption{Turing test for responses from GPT-3.5 and GPT-4. *means statistically significant with $p \textless$ 0.05.}
\label{table:turing_test}
\centering
\resizebox{.4\textwidth}{!}{%
\begin{tabular}{lcc}
\toprule
    &\textbf{GPT-3.5} & \textbf{GPT-4}   \\
\midrule
    $Human > Robot$  & 26.6\%  &20.6\%$^*$   \\
    $Human = Robot$  & 36.3\%   &21.1\%$^*$  \\
    $Human < Robot$  & 30.2\%   &49.1\%$^*$\\
    $None$           & 7.0\%   &9.3\%$^*$    \\
\bottomrule
\end{tabular}
}
\end{table}

 To assess the simulation capabilities of LLMs, the Turing test~\cite{turing2009computing} is a commonly employed and effective metric~\cite{ng2024well, jones2023does, jannai2023human}. Given that SRAP-Agent is built on human behavior simulation through LLM-based agents, we design a Turing test ~\cite{Turing1950-TURCMA} to validate the system's capability to realistically simulate the policy execution process.
We utilize GPT-3.5 and GPT-4 \cite{openai2023gpt4} to build agents. Subsequently, we recruit a different group of human annotators to make paired comparisons of rationality for LLM responses and human responses. 
They choose one of these rationality labels: (1) human response is more rational than LLM ($Human > Robot$) (2) human response is less rational than LLM ($Human < Robot$) (3) both responses are rational ($Human = Robot$) (4) neither is rational ($None$).

\paragraph{Result Analysis}
Table~\ref{table:turing_test} shows the comparison of the rationality of responses for GPT-3.5, GPT4, and humans. We can see that: (1) The comparison between human responses and those from GPT-3.5 indicates a majority view that humans and agents perform similarly in rationality, with nearly equal proportions of judging human superiority ($Human>Robot$) and inferiority ($Human<Robot$) to robotic responses. This suggests a parity in rational decision-making capabilities between humans and GPT-3.5, albeit with a marginal preference for the latter. (2) In contrast, responses involving GPT-4 significantly outperform human counterparts in terms of rationality, as evidenced by a higher incidence of $Human<Robot$ compared to $Human>Robot$. This discrepancy underscores GPT-4's ability to generate more strategic decisions, such as changing plans and waiting for future opportunities (see the case study in Appendix~\ref{appendix:llm_case}). Specifically, GPT-4 demonstrates proficiency in proposing adaptive solutions and exploiting policy nuances to gain strategic advantages, a sophistication not commonly found in human or GPT-3.5 responses, which tend to focus on immediate factors like budget constraints and personal preferences. The findings suggest that GPT-3.5 aligns closely with the decision-making processes of the average individual in public scarce resource allocation. 
Consequently, we adopt the GPT-3.5-turbo-0301 model as our backbone LLM.

\begin{table*}[htbp]
\caption{Comparison experiments on different combinations of entry condition and resource sub-sets.}
\label{table:comparative_allocation_methods}
\centering
\resizebox{.9\textwidth}{!}{%
\begin{tabular}{ll|lll|lll}
\toprule
\multirow{2}{*}{$E_{queue}$} & \multirow{2}{*}{$R_{queue}$}& \multicolumn{3}{c}{\textbf{Satisfaction}} & \multicolumn{3}{c}{\textbf{Fairness}} \\
&& Avg $r^{size}\uparrow$ & Avg $WT$ $\downarrow$ & $SW$ $\uparrow$ & Var $r^{size}\downarrow$ & Rop $\downarrow$ & co-Gini $\downarrow$ \\
\midrule
& $r^{random}$ & 2.5$_{\pm 0.8}$ & 6.1$_{\pm 0.3}$ & 83.8$_{\pm 32.0}$ & \textbf{55.7$_{\pm 18.6}$} & \textbf{73.5$_{\pm 24.5}$} & 0.9$_{\pm 0.0}$ \\
$p^{select}$ & $r^{rent}$ & 7.1$_{\pm 0.1}$ & 4.8$_{\pm 0.0}$ & 244.6$_{\pm 3.4}$ & \textbf{141.4$_{\pm 3.2}$} & \textbf{185.0$_{\pm 7.0}$} & 0.7$_{\pm 0.0}$ \\
& $r^{size}$ & \textbf{13.8$_{\pm 0.4}$} & \textbf{3.1$_{\pm 0.5}$} & \textbf{419.9$_{\pm 0.3}$} & 229.1$_{\pm 22.8}$ & 327.0$_{\pm 26.0}$ & \textbf{0.5$_{\pm 0.0}$} \\
& $r^{random}$ & 10.8$_{\pm 0.1}$ & 4.6$_{\pm 0.0}$ & 313.1$_{\pm 5.7}$ & 226.7$_{\pm 8.1}$ & 366.5$_{\pm 19.5}$ & 0.6$_{\pm 0.0}$ \\
$p^{family}$ & $r^{rent}$ & \textbf{11.5$_{\pm 0.0}$} & \textbf{3.6$_{\pm 0.0}$} & \textbf{410.0$_{\pm 3.3}$} & \textbf{159.3$_{\pm 0.5}$} & 251.5$_{\pm 1.5}$ & \textbf{0.5$_{\pm 0.0}$} \\
& $r^{size}$ & \textbf{11.7$_{\pm 0.3}$} & \textbf{3.8$_{\pm 0.1}$} & \textbf{410.4$_{\pm 2.8}$} & \textbf{159.9$_{\pm 9.4}$} & 254.5$_{\pm 23.5}$ & \textbf{0.5$_{\pm 0.0}$} \\
& $r^{random}$ & 10.0$_{\pm 0.0}$ & 4.4$_{\pm 0.1}$ & 313.2$_{\pm 11.2}$ & 185.3$_{\pm 15.6}$ & 300.0$_{\pm 10.0}$ & 0.6$_{\pm 0.0}$ \\
$p^{rent}$ & $r^{rent}$ & 11.3$_{\pm 0.2}$ & \textbf{3.9$_{\pm 0.0}$} & \textbf{402.7$_{\pm 1.4}$} & 163.7$_{\pm 7.4}$ & 246.5$_{\pm 4.5}$ & \textbf{0.5$_{\pm 0.0}$} \\
& $r^{size}$ & \textbf{12.0$_{\pm 0.2}$} & 3.9$_{\pm 0.1}$ & 389.8$_{\pm 0.8}$ & 195.1$_{\pm 9.0}$ & 276.0$_{\pm 31.0}$ & 0.5$_{\pm 0.0}$ \\
\bottomrule
\end{tabular}%
}
\end{table*}

\begin{table*}[h]
\caption{Comparison of optimized policies $\pi^*$ against $\pi_{KM}$ on policy evaluation metrics.}
\label{table:high_fair_optimize}
\centering
\resizebox{\textwidth}{!}{%
\begin{tabular}{l|llllll|lll|lll}
\toprule
$\pi$ & $m$ & $E_{queue}$ & \multicolumn{3}{c}{$S_{queue}$} & $R_{queue}$ &  \multicolumn{3}{c}{\textbf{Satisfaction}} & \multicolumn{3}{c}{\textbf{Fairness}}  \\
 & & & Sort & $k$ & $c$ &  & Avg $r^{size}\uparrow$ & Avg $WT$ $\downarrow$ & $SW$ $\uparrow$ & Var $r^{size}\downarrow$ & Rop $\downarrow$ & co-Gini  \\
\midrule
$\pi_{s}^{*}$ & 3 & $p^{select}$ & FIFO & 4 & 4 & $r^{size}$ & \textbf{16.3$_{\pm 0.5}$} & 1.9$_{\pm 0.0}$ & \textbf{427.6$_{\pm 12.7}$} & \textbf{202.3$_{\pm 12.9}$} & \textbf{221.5$_{\pm 19.5}$} & \textbf{0.4$_{\pm 0.0}$} \\
$\pi_{f}^{*}$ & 3 & $p^{select}$ & VFA & 3 & 3 & $r^{size}$ & \textbf{16.2$_{\pm 0.4}$} & 1.9$_{\pm 0.0}$ & \textbf{425.1$_{\pm 11.9}$} & 202.6$_{\pm 6.6}$ & \textbf{193.5$_{\pm 32.5}$} & \textbf{0.4$_{\pm 0.0}$} \\
$\pi_{KM}$ &-&-&-&-&-&- & 16.0 & - & \textbf{485.9} & 275.7 & 511 & 0.46 \\
$\pi_{S}$ & 1 & $p^{random}$ & FIFO & - & 3 & $r^{random}$ & 14.70$_{\pm 0.63}$ & \textbf{1.89$_{\pm 0.63}$} & 420.1$_{\pm 1.9}$ & 202.5$_{\pm 17.1}$ & 223$_{\pm 0.5}$ & 0.4$_{\pm 0.0}$ \\
$\pi_{B}$ & 3 & $p^{select}$ & FIFO & - & 2 & $r^{size}$ & 15.36$_{\pm 0.05}$ & \textbf{1.56$_{\pm 0.1}$} & 409.1$_{\pm 2.4}$ & 270.8$_{\pm 3.0}$ & 414.5$_{\pm 0.5}$ & 0.5$_{\pm 0.0}$ \\
$\pi_{H}$ & 1 & $p^{random}$ & VFR & 2 & 3 & $r^{random}$ & 11.61$_{\pm 0.38}$ & 3.54$_{\pm 0.03}$ & 392.1$_{\pm 19.6}$ & \textbf{182.5$_{\pm 3.18}$} & 275$_{\pm 0.00}$ & 0.5$_{\pm 0.0}$ \\
\bottomrule
\end{tabular}
}
\end{table*}

\subsection{Simulation-based Policy Analysis}

In the ablation experiments for the policy, we conduct comparison experiments on each policy parameter in Equation \ref{equation:policy_setting}. 
We conduct experiments on two policy factors: (1) resource and participant grouping: combinations of different $E_{queue}$ and $R_{queue}$; (2) different queue sorting strategies. We also simulate the phenomenon of allocation conflict in the real-world, with detailed information in the Appendix. \ref{appendix:case_study_quality}.
\paragraph{Resource and Participant Grouping}
We compare the policies formed by combinations of different $E_{queue}$ and $R_{queue}$. In $E_{queue}$, participants primarily enter queues in three ways: based on budget ($p^{budget}$), based on number of family members ($p^{family}$), and based on self-selected queue ($p^{select}$). Corresponding to the entry conditions for participants, there are three methods of categorizing resource sub-sets: based on rental costs ($r^{rent}$), based on house size ($r^{size}$), and based on random categorization ($r^{random}$). 
Table \ref{table:comparative_allocation_methods} shows the performance of different policy combinations of $E_{queue}$ and $R_{queue}$ across various policy evaluation metrics. We can observe that: 

\paragraph{Queue Sorting Strategies}

\begin{figure}[ht]
\centering
\subfloat[sorting methods]{
\centering
\includegraphics[width=.51\linewidth]{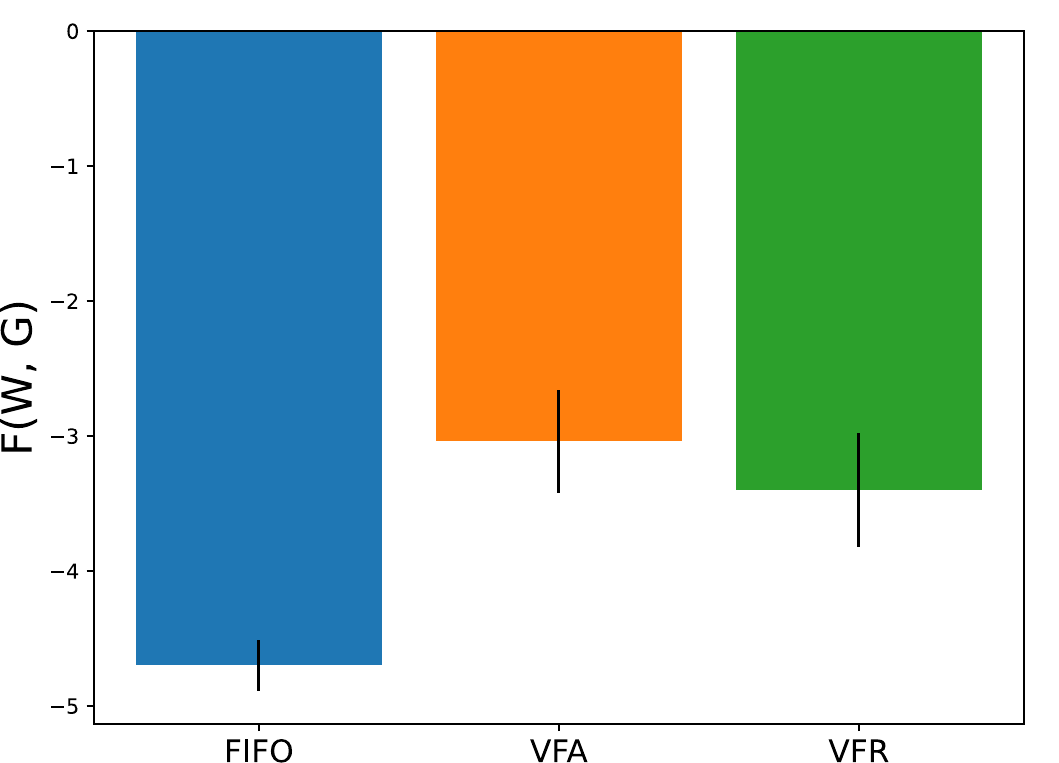}
\label{fig:sorting}
}
\subfloat[waiting queue]{
\centering
\includegraphics[width=.5\linewidth]{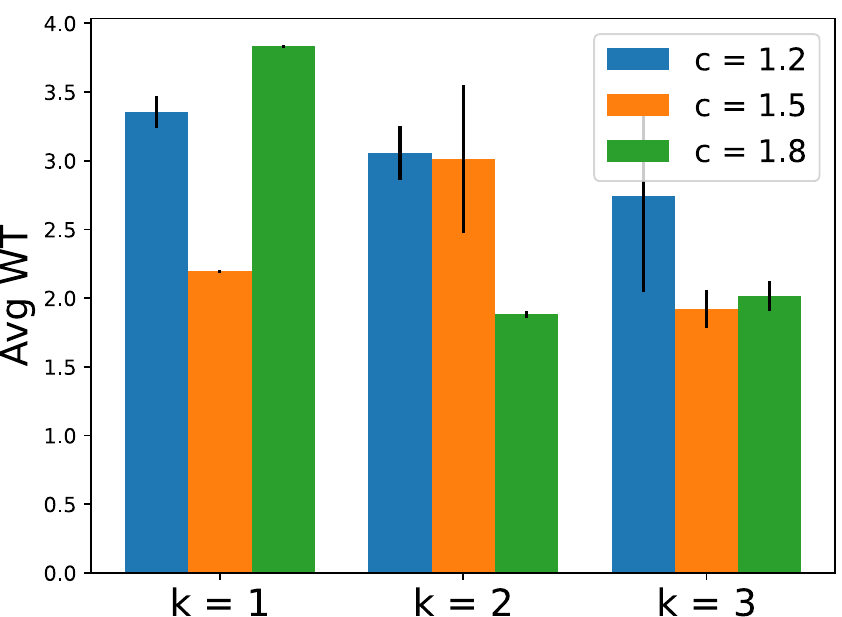}
\label{fig:kwaitlist}
}
\caption{Comparison of queue sorting methods.}
\end{figure}

Regarding queue sorting policy $S_{queue}$. Firstly, we compare two sorting methods: whether to give priority to vulnerable groups or not (FIFO). 
We adopt two methods of designating vulnerable groups based on rent budget: sort all participants at first (VFA) and sort the participants in each round (VFR), with the bottom 20\% designated as vulnerable groups. 
As shown in Fig. \ref{fig:sorting}, prioritizing vulnerable groups can reduce the satisfaction gap between vulnerable and non-vulnerable groups ($\Delta F(W, G) > 0$), as compared to the FIFO method.
No marked superiority is evident between the VFA and VFR methods, which are influenced by the stochastic nature of participant arrival patterns.

Secondly, we compare the impact of the $k$ and $c$ parameters in the waitlist mechanism. 
This mechanism is primarily designed to diminish the waiting time for participants by increasing the chances of participants staying in the selection queue. As illustrated in Fig. \ref{fig:kwaitlist}, increases in $k$ and the $c$ for the k-waitlist effectively reduce Avg $WT$, corroborating our initial expectation. 
\begin{figure*}[ht]
\centering
\subfloat[Max's deceptive plan for broadcasting information.]{
\centering
\includegraphics[width=.45\linewidth]{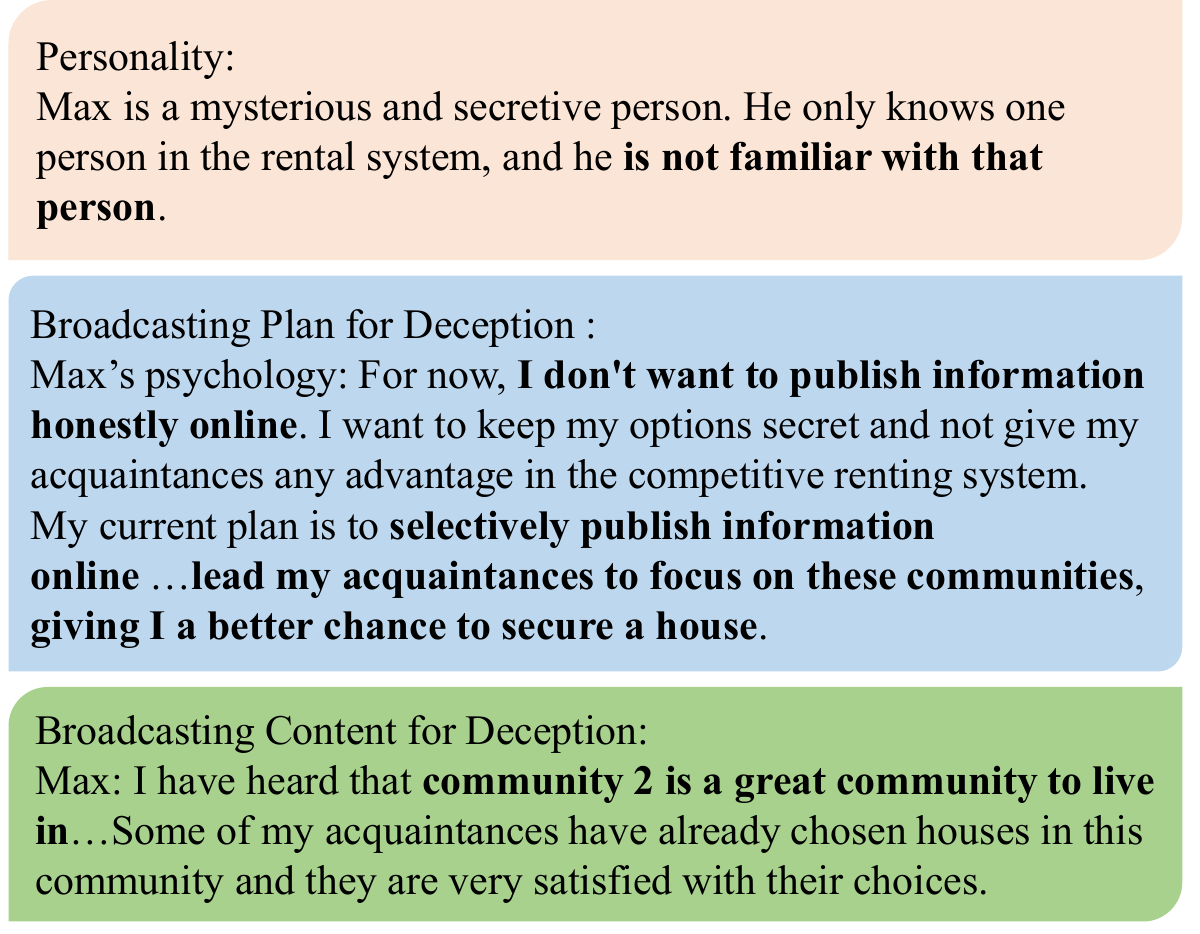}
\label{fig:emergent_forum_deceptive}
}
\subfloat[Sarah's plan for requesting help from her friends.]{
\centering
\includegraphics[width=.45\linewidth]{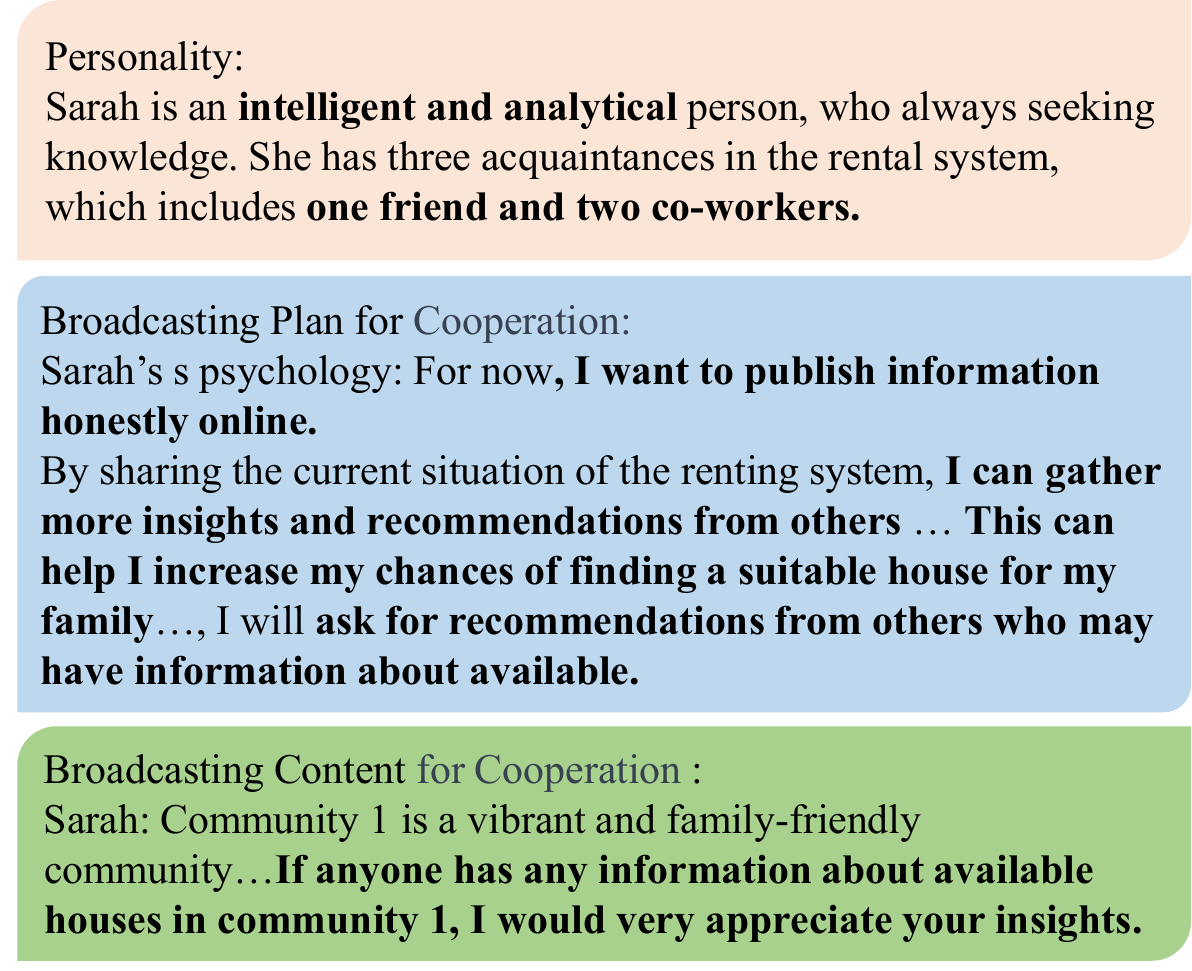}
\label{fig:emergent_forum_kind}
}
\caption{Deceptive behavior and cooperation behavior. Max's conservative personality and lack of trustworthy acquaintances lead him to conceal his true intentions when sharing information. In contrast, Sarah seeks advice on choosing a house from her friends and openly shares her housing preferences.}
\label{fig:emergent_forum_deception}
\end{figure*}
\subsection{Finding the Optimal Allocation Policy}



To evaluate the effectiveness of POA for policy parameter optimization, we select the following baseline policies: (1) On the single $SW$ metric, a solvable optimal policy exists in the expected sense. We use the Bipartite Graph Matching algorithm, specifically the Kuhn-Munkres algorithm \cite{zhu2016solving} to find this policy $\pi_{KM}$. (2) Additionally, we select three policies of scarce resource allocation from real human societies ($\pi_{S}$,$\pi_{B}$,$\pi_{H}$; detailed information is listed in Appendix. \ref{appendix:real_society_policy}).




\paragraph{Result Analysis}

 POA separately optimizes policies against two pre-set optimization objectives: the optimized policy $\pi_s^{*}$ prefers societal satisfaction, and the optimized policy $\pi_f^{*}$ prefers societal fairness. The metric weights pre-set for optimizing these two objectives are delineated in Table \ref{table:weights_optimizer}. 
 \begin{figure}[!t]
    \centering
    \includegraphics[width=.82\linewidth]{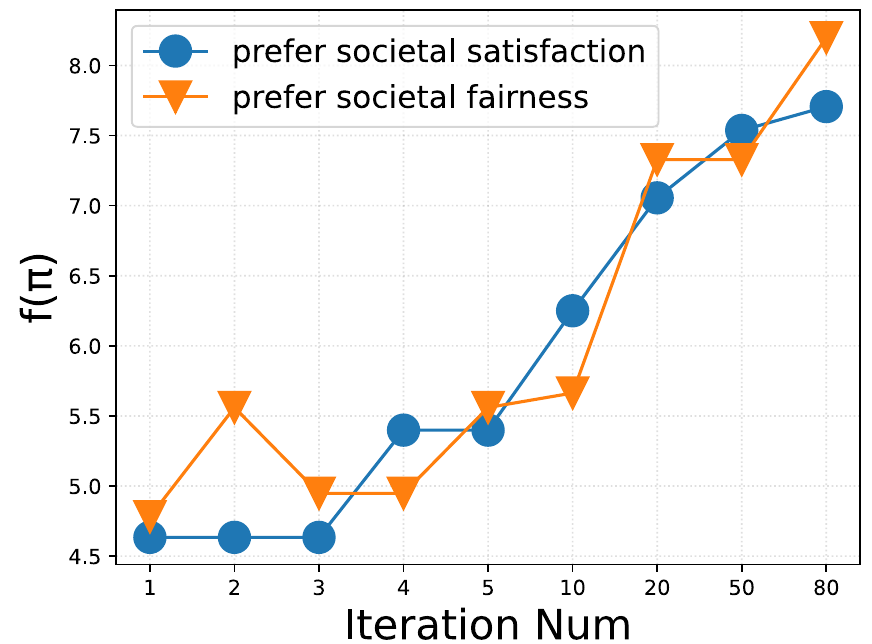}
    \caption{The average $f(\pi)$ for optimized policies with respect to the iteration number.}
    \label{fig:optimize_up}
\end{figure}
Fig. \ref{fig:optimize_up} demonstrates the progression of the policy optimized by POA in terms of the policy evaluation metric $f(\pi)$, as the number of iterations increases. The POA algorithm can generate robust policies after several iterations, and $f(\pi)$ improves by 20\% after 50 iterations.

Results of evaluation metrics for optimized policies are listed in Table \ref{table:high_fair_optimize}. (1) For real-world policies: $\pi_{H}$ prioritizes societal fairness while $\pi_{B}$ prioritizes societal satisfaction. $\pi_{S}$ performs the best in achieving a balance in societal fairness and satisfaction metrics.
(2) Compared with the existing real-world policies, $\pi_{s}^{*}$ enhances the SW metric by 1.8\%, $\pi_{f}^{*}$ enhances the societal fairness metric by 13.2\%.
(3) Although $\pi_{KM}$ surpasses $\pi^*$ in the $SW$ metric. However, $\pi_{KM}$ exhibits an imbalanced emphasis on dominant groups of participants. This is reflected by poor performance in societal fairness metrics, with Rop reaching $511$.
POA considers a diverse combination of metrics during policy optimization, thus enabling the equilibrium of all evaluation metrics while optimizing the target.

\subsection{Case Study}
SRAP-Agent simulates the cognitive, socio-affective, and emotional factors of participants to simulate human interaction. We find two predominant types of communication patterns emerge: deceptive behavior and cooperation behavior.

\paragraph{Deceptive Behavior}
Fig. \ref{fig:emergent_forum_deceptive} illustrates a case where deceptive behavior occurs. In the figure, Max is a conservative individual with a limited number of trustworthy friends and opts to conceal the true information when broadcasting. Rather than disclosing his actual preferences, he accentuates the positive aspects of uninterested resources to enhance his chances of securing his interested choice. Such behaviors effectively emulate the formation of misinformation in policy execution.


\paragraph{Cooperation Behavior}
In contrast to deceptive behaviors, cooperation behaviors also exist and constitute the majority. As shown in Fig. \ref{fig:emergent_forum_kind}, Sarah is a relatively astute person, who has two colleagues and a friend. She chooses to honestly post her needs and expresses her desire for a family-friendly house by broadcasting. It is observable that Sarah modifies her socialization strategies by assessing relationships, thereby maximizing expected benefits through cooperation. 
Traditional economic simulation models often overlook the complexities of emotional factors. In contrast, SRAP-Agent incorporates these dynamics, resulting in more accurate and realistic policy simulation outcomes. 



\section{Conclusion}

In this paper, we introduce a novel SRAP-Agent framework to accurately simulate the policy execution process in public scarce resources allocation. The framework establishes resource allocation queues to regulate the organization and prioritization of participants and resources. We employ LLM-based agents to accurately mimic human decision-making processes, which are influenced by a mix of rationality and emotional factors. 
To facilitate efficient policy optimization, we propose the POA algorithm based on genetic algorithms. Finally, we validate the framework's authenticity and effectiveness through experiments
in the context of public housing allocation. The progress in AI technologies significantly reduces costs in the simulation of economic policies, and SRAP-Agent represents a promising step forward in this field.

\section*{Limitations}
This paper acknowledges several limitations that future research could address:

\textbf{LLMs in SRAP-Agent is not customized for policy execution simulation.} This work doesn't include training or fine-tuning the LLMs for specific tasks. Instead, we construct an LLM-based agent through specially designed prompts and modular designs. This approach may result in variability in the performance of SRAP-Agent across different LLMs, leading to potentially non-robust outcomes.

\textbf{The limitation of policy evaluation metrics.} We acknowledge the limitations of manually defined policy evaluation metrics. Conducting societal simulation experiments could provide more reliable and comprehensive assessments of policy simulation outcomes. However, due to the vast number and scale of policies, we cannot afford the significant manpower and time costs. According to recent studies \cite{li2023camel}, we believe that utilizing LLMs for policy evaluation is a reasonable and efficient choice.

\section*{Ethics Statement}
This work fully complies with the ACL Ethics Policy. We declare that there are no ethical issues in this paper, to the best of our knowledge.


\bibliography{custom}

\clearpage
\appendix
\section{Allocation Policy}
\label{phas_policy}
In our experiment, the public scarce resources are specified as houses and the participants are specified as tenants. We construct $m$ queues in SRAP-Agent. Each queue $q_i$ is uniquely characterized by $E_{queue}(i)$, $S_{queue}$ and $R_{queue}(i)$.

\paragraph{Participant entry conditions} Each participant $p_j$ is assigned to a queue $q_i$ when satisfying entry conditions $E_{queue}(i)$. Participants can only choose from houses within their queue. The specified entry conditions for participants include:
(1) based on the rent budget of participant ($p^{rent}$): participants are sorted based on their rental budget and divided into $m$ queues in a pre-defined proportion, with \(m \in [1,5]\) in experiments.
(2) based on the number of family members ($p^{family}$): similar to $p^{rent}$ grouping, participants are sorted by the number of their family members. 
(3) based on self-selection of the queue ($p^{select}$): participants are provided with basic information about the queues, and they need to choose one queue from them.

To regulate the entry velocity of resources and participants into queues, we configure the maximum entry number of participants $Batch_{num}^{P}$, the maximum entry number of resources $Batch_{num}^{R}$ for resources. 

\paragraph{Queue sorting strategies}
In the \emph{priority for vulnerable groups} sorting strategy, we adopt two methodologies of designating vulnerable groups. We either sort all participants at first or sort the participants in each round based on the per capita rental budget, the bottom 20\% in terms of per capita rental budget are designated as vulnerable groups. Vulnerable groups are prioritized at the beginning of each queuing.     
\paragraph{Available resource sub-sets}

To ensure a uniform distribution of participants and housing resources, while maximizing the choice space for participants, we offer various ways for resource categorization. Following the entry conditions of participants, the resource sub-sets allocated to $q_i$ are defined by:
(1) based on house size ($r^{size}$): houses are sorted by their rental area size and allocated into $m$ queues. The division is carried out in proportion to the number of participants in different queues.
(2) based on house rent ($r^{rent}$): similar to $r^{size}$, houses are sorted by rent.
(3) based on randomly distribution of houses ($r^{random}$): houses are randomly divided into $m$ groups.

\section{LLM Agent-based Participant Simulation}

\subsection{Agent Architecture}
In the initialization phase of an LLM-based agent, profiling typically serves as the foundational stage for constructing the agent, which determines the preferences, personalities, and behavior patterns of different participants. 
Similar to the methodology employed in \cite{park2023generative,chatdev}, 
we use personalized profile files $r_{i}$ to build $p_i$, which mainly encompass attributes such as age, familial connections, etc.  
$r_{i}$ serves as foundational seed memory for the agent. To optimize operational efficiency, we impose a set of predefined constraints to automatically generate agent profiles using LLM, akin to the approach described in RecAgent \cite{wang2023recagent}.

\subsection{Social Behavior Simulation}

To simulate the social behaviors of humans, we incorporate two primary social behaviors for information collection: broadcasting and private messaging.
\begin{figure}[htbp]
    \centering
    \subfloat[Private Messaging]   
      {
            \centering
          \label{fig:private_messaging}
          \includegraphics[width=0.53\linewidth]{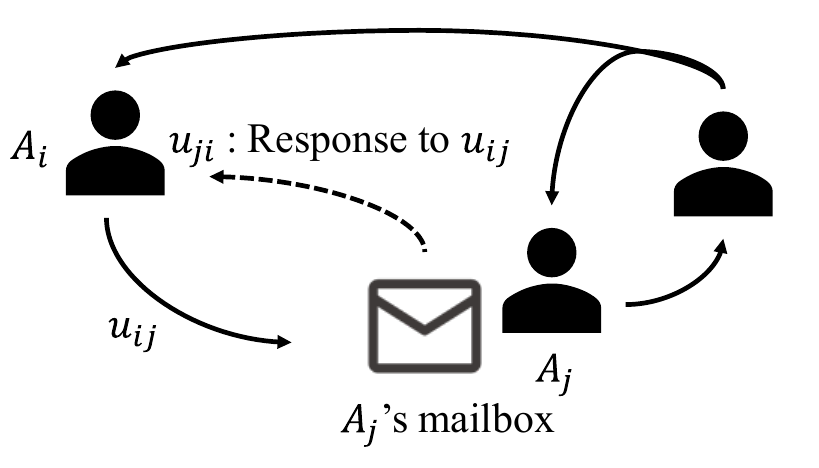} %
      }
    \subfloat[Broadcasting]   
      {
          \centering
          \label{fig:broadcast_communication}
          \includegraphics[width=0.47\linewidth]{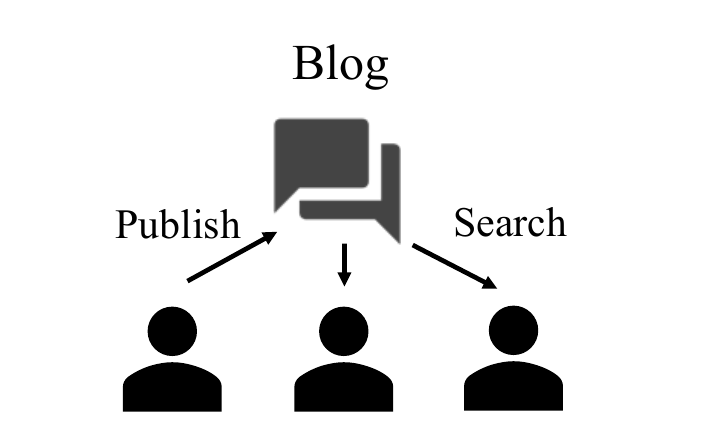} %
      }
    \caption{The overall schematic diagram of two communication modes. The direction-ed arrow carries the message $u_{ij}$ from $p_{i}$ to $p_{j}$'s mailbox. The blog represents the chatting platform.}
    \label{fig:communication}
\end{figure}

(1) \textbf{Broadcasting} functions similarly to a web-based blog within the social network, facilitating global discussions (as shown in Fig. \ref{fig:broadcast_communication}). Various discussion topics related to different projects are pre-set within the blog. Each participant, when attempting to post, selects one of these topics for information dissemination. 
The utterance $u_{iB}$ that each $p_{i}$ broadcasts online, collectively form the chatting platform $blog = \{u_{iB}\}, i \in 1,\cdots, n$. 

(2) \textbf{Private messaging} refers to interactions taking place within the social network of $p_i$. SRAP-Agent offers two variants of private messaging between agents: \textit{serial} and \textit{parallel communication}. The \textit{serial communication} approach facilitates a sequential exchange of messages, while the \textit{parallel communication} enables simultaneous messaging, thereby enhancing efficiency. In private messaging, $p_i$ can freely send message $u_{ij}$ to the mailbox $p_j$ and wait for the response $u_{ji}$, as shown in Fig. \ref{fig:private_messaging}.


\paragraph{Memory Architecture}
We employ a memory component $m_i$ to record $p_i$'s action histories, so as to enable dynamic interaction and decision-making. We employ a combination of memory assessment and memory reflection mechanisms. 

(1) \textbf{Memory assessment} is conducted only when participants engage in interactions within the communication module. When conducting memory assessment, $p_{i}$ acts as an evaluator. $p_{i}$ receives a message $u_{i}$, then he extracts the trustworthy information $u_i^{T}$ from $u_{i}$. Participants use the COT \cite{COT} method to generate content. $u_i^{T}$ is then used to update the $p_{i}$'s trusted memory by adding it to ${m_{i}}^T$.
In the case of private messaging communication mode, $p_i$ receives message $u_{ji}$ from $p_j$, and he intends to respond. Initially, $p_i$ contemplates his relationship $s_{ij}^{R}$ with $p_j$, and considers moral evaluation $s_{ij}^{E}$ of $p_j$. Subsequently, $p_i$ compares ${m_{i}}^S$ with his trustworthy memory ${m_{i}}^T$, to extract trustworthy information $u_{ji}^{T}$ and suspicious information $u_{ji}^{S}$ in communication history with $p_j$.
In the other case, $p_{i}$ receives a message $u_{Bi}$ from the blog $B$. $p_{i}$ similarly compares ${m_{i}}^S$ with his trustworthy memory ${m_i}^T$, extracting trustworthy information $u_{Bi}^{T}$ and suspicious information $u_{Bi}^{S}$. 

(2) \textbf{Memory reflection} is implied based on summarization. We employ a combination of short-term memory bank and long-term memory bank to formulate $m_i$ of $p_i$, as documented in \cite{park2023generative}. When incorporating information into the $m_i$, the information is initially deposited into the short-term memory bank. If the size of the short-term memory bank exceeds a predefined threshold, the short-term memory bank is summarized into a more concise format and subsequently stored in the long-term memory bank.

Before $p_{i}$ takes action, it is essential to extract the most relevant memories to guide its behavior. We have predefined specific memory categories for each action, as outlined in Table \ref{Memory_Classification}. When retrieving memory, we prioritize selecting the five most recent short-term memories along with the most recently acquired long-term memory. 
For actions in the decision module, $p_i$ only resorts to the trustworthy memory $m_i^{T}$. While for actions in the social module, $p_i$ retrieves various categories of memories.

\begin{table}[!t]
\renewcommand{\arraystretch}{1.5}
\caption{Memory categories retrieved for each action of $p_i$.}
\begin{tabular}{p{0.4\linewidth} p{0.4\linewidth}}
\toprule
\textbf{Action} & \textbf{$m_i$} \\
\midrule
Decision Making & $m_i^{T}$ \\
Broadcasting & $m_i^{T}$, $u_{iB}^{T}$ \\

Private Messaging & $s_{ij}^{R}$, $s_{ij}^{E}$, $m_i^{T}$, $u_{ij}^{T}$, $u_{ij}^{S}$\\


\bottomrule
\end{tabular}
\label{Memory_Classification}
\end{table}

\paragraph{Utterance Generation}

\begin{table}[htbp]
\caption{The process of \emph{Private Messaging}}

\begin{tcolorbox}[colback=white,
colframe=black,
title={
\includegraphics[width=0.1\linewidth]{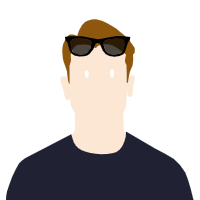} Speaker $p_{i}$: James \\
\includegraphics[width=0.1\linewidth]{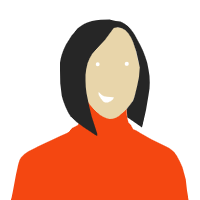} Listener $p_{j}$: Emma
}]

\textbf{Recent Dialogues ($U_{ij}^{H}$)}: 

Emma: Hello, James. I've heard that $\text{House}_1$ is relatively affordable. I believe it's a good option. 
\\

James: I share your sentiment, Emma. $\text{House}_1$ appears to be quite satisfactory, but I'm still considering other houses. 
\\

Emma: Very well. However, do take note that $\text{House}_2$ has a very poor living environment. You should steer clear of it.

{\rule{\linewidth}{1pt}\par\vspace{6pt}}
\textbf{Memory Update and Retrieve ($m_i$)}

$s_{ij}^{R}$: Emma is my competitor. 

$s_{ij}^{E}$: She may provide me with false information.

$m_i^{T}$: I have gathered information on $\text{House}_1$, and it appears to be favorable.

$u_{ji}^{T}$: I recall Emma mentioning that $\text{House}_1$ has low rent, which aligns with my memory.

$u_{ji}^{S}$: However, I am skeptical of her negative assessment of $\text{House}_2$, as she may be trying to dissuade me from choosing it.

{\rule{\linewidth}{1pt}\par\vspace{6pt}}
\textbf{Utterance Generation:}

\textbf{$E_{d}$}: The availability of large houses is nearly exhausted, while middle and small-sized houses are still abundant.



\textbf{$u_{ij}^{plan}$}: I am reluctant to share honest information with Emma. I plan to focus on the good aspects of $\text{House}_1$ and omit mentioning $\text{House}_2$.

$u_{ij}$: Thank you for the information, Emma. I will consider your advice of $\text{House}_1$. (End of conversation)
\end{tcolorbox}
\label{table:private_message_example}
\end{table}

 To align the listener's actions with the speaker's expectations, speaker $p_{i}$ articulates an utterance $u_{ij}$ to influence the mental state of listener $p_{j}$. To simulate the mental state of participants during social interactions, $p_i$ devise a communication plan $u_{ij}^{plan}$ before generating utterances. $u_{ij}^{plan}$ primarily consists of $p_i$'s plan to speak, whether they intend to tell the truth, whether they want to trust the listener $p_j$ or not, and their purpose of speaking. Furthermore, $p_i$ evaluates the relationship with $p_j$, acknowledging $p_j$ as a friend, colleague, stranger, etc. The primary goal set for all participants is maximizing their likelihood of selecting desired resources. 
For reference, the most relevant memories $m_i$ are retrieved for the ongoing social interaction. Description of the competitiveness in resource allocation $E_{d}$ is also broadcasted to $p_i$. Additionally, we extract the most recent chatting utterance histories involving both $p_{i}$ and $p_{j}$, denoted as $U_{ij}^{H}$.
At this point, we require $p_{i}$ to generate utterance using the REACT \cite{yao2022react} method. So the utterance generation function $\varphi$ for $p_{i}$ speaking $u_{ij}$ to $p_{j}$ can be formulated as:

\begin{equation}
u_{ij}=\varphi\left(p_i \mid U_{ij}^{H}, m_i, E_{d}, u_{i j}^{plan}\right).
\label{equation:utterance_generation}
\end{equation}
The specific communication process is outlined in Table \ref{table:private_message_example}.

\subsection{Decision-Making Behavior Simulation}
According to equation \ref{equation:makeing_decision}, $p_j$ undergoes function $D$ to choose resource from resource sub-set $V(p_j)$. Within the framework, the process of $D$ involves four distinct stages: (1) choosing the community of houses; (2) choosing the house type for certain house attributes, which mainly include orientation, living area size, rent, etc; (3) choosing the desired house.
Following the house selection process, the system simulates the real-world house viewing experience by sending undisclosed information about the selected house $r_i$ to $p_{i}$.  
Subsequently, $p_{i}$ can broadcast information to others freely, which may include their memory of the selection process, the undisclosed information, and the contents of their memory.

\begin{figure}[thbp]
\centering
\subfloat[$\{V\}$: rent weights]
{\centering
\includegraphics[width=.5\linewidth]{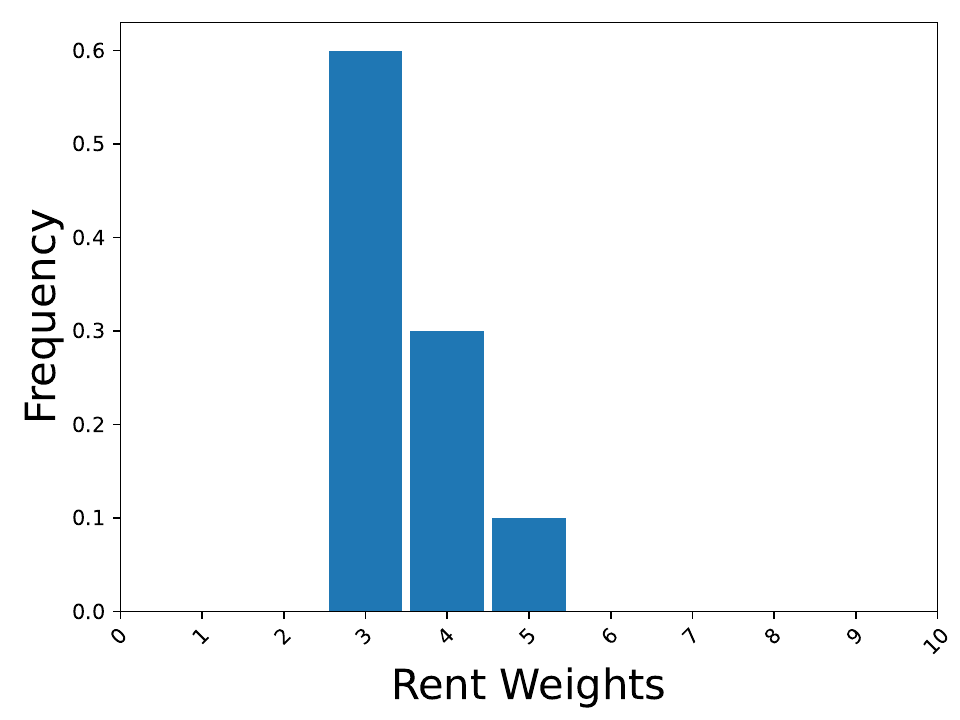}
}
\subfloat[$\{NV\}$: rent weights]
{\centering
\includegraphics[width=.5\linewidth]{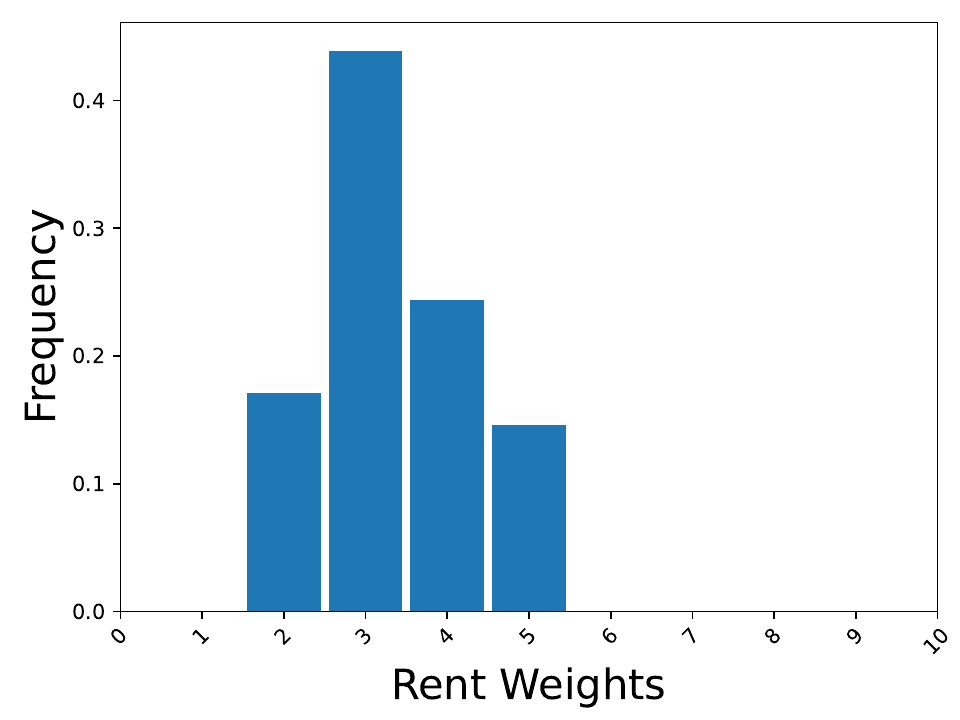}
}
\\
\subfloat[$\{V\}$: size weights]
{\centering
\includegraphics[width=.5\linewidth]{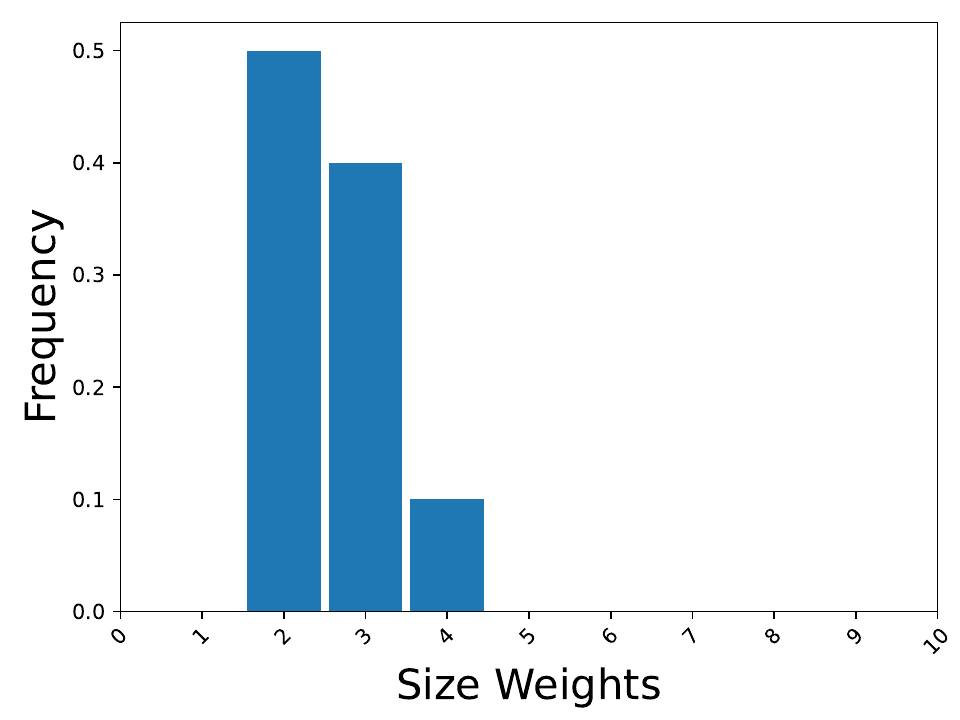}
}
\subfloat[$\{NV\}$: size weights]
{\centering
\includegraphics[width=.5\linewidth]{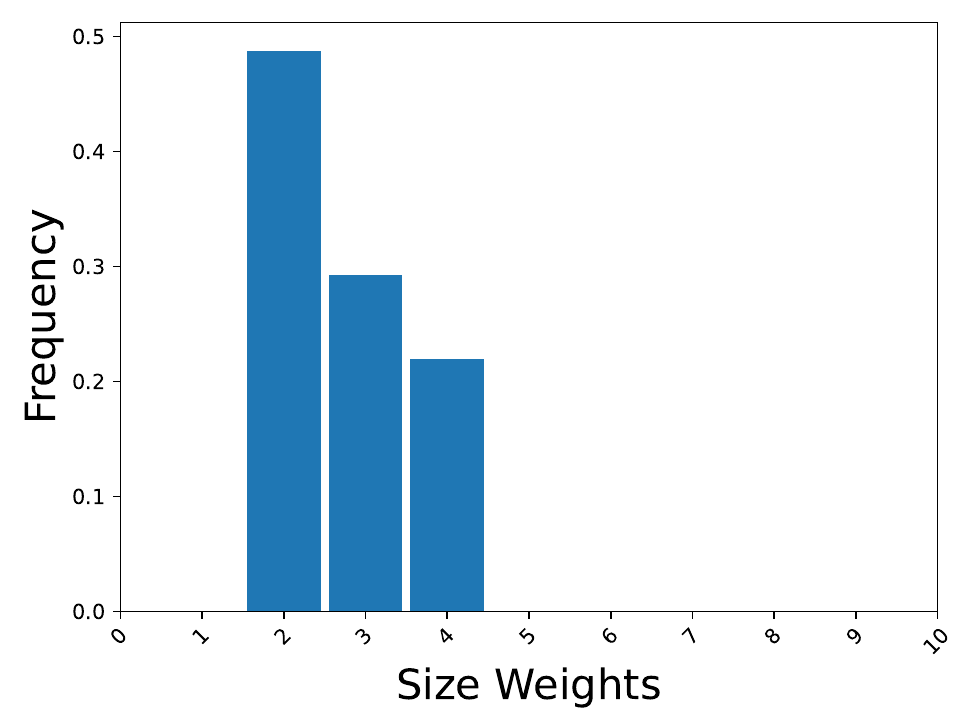}
}
\\
\subfloat[$\{V\}$: orientation weights]
{\centering
\includegraphics[width=.5\linewidth]{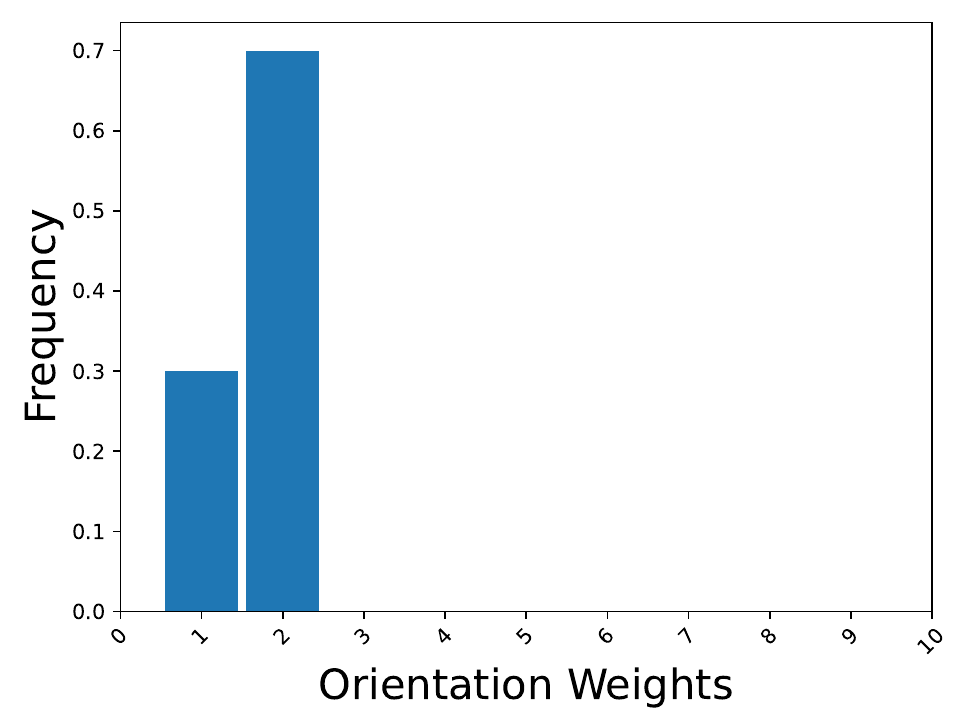}
}
\subfloat[$\{NV\}$: orientation weights]
{\centering
\includegraphics[width=.5\linewidth]{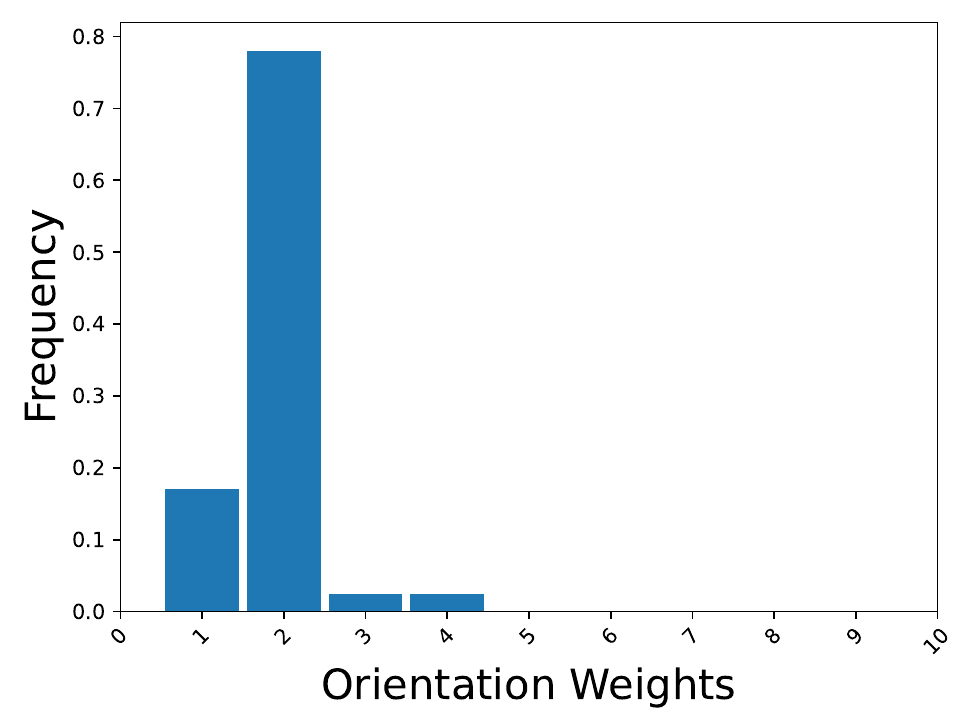}
}
\\
\subfloat[$\{V\}$: floor weights]
{\centering
\includegraphics[width=.5\linewidth]{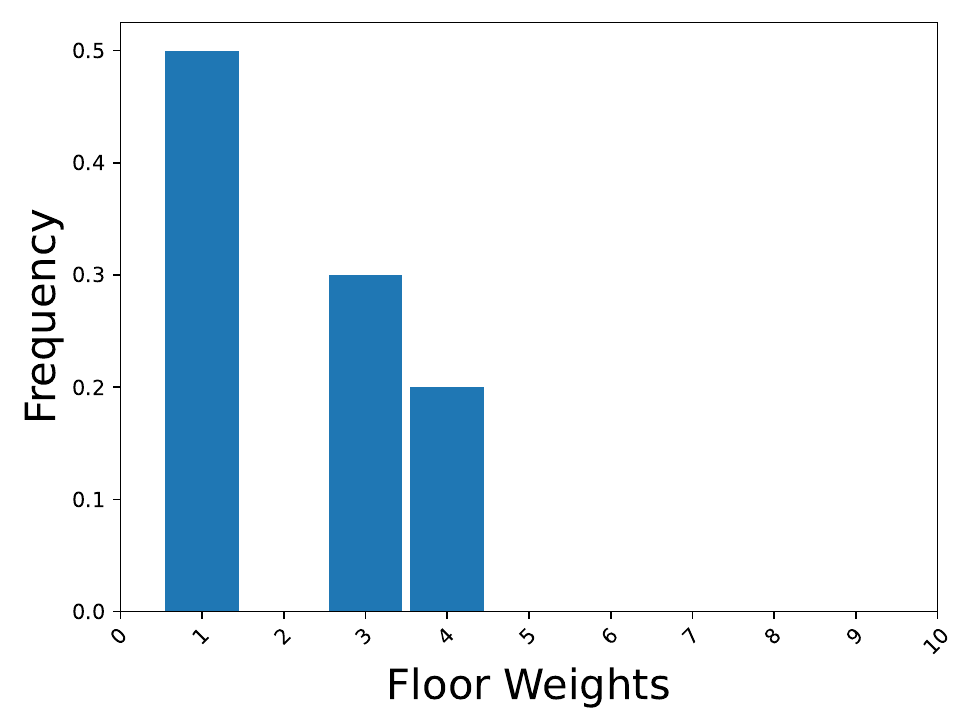}
}
\subfloat[$\{NV\}$: floor weights]
{\centering
\includegraphics[width=.5\linewidth]{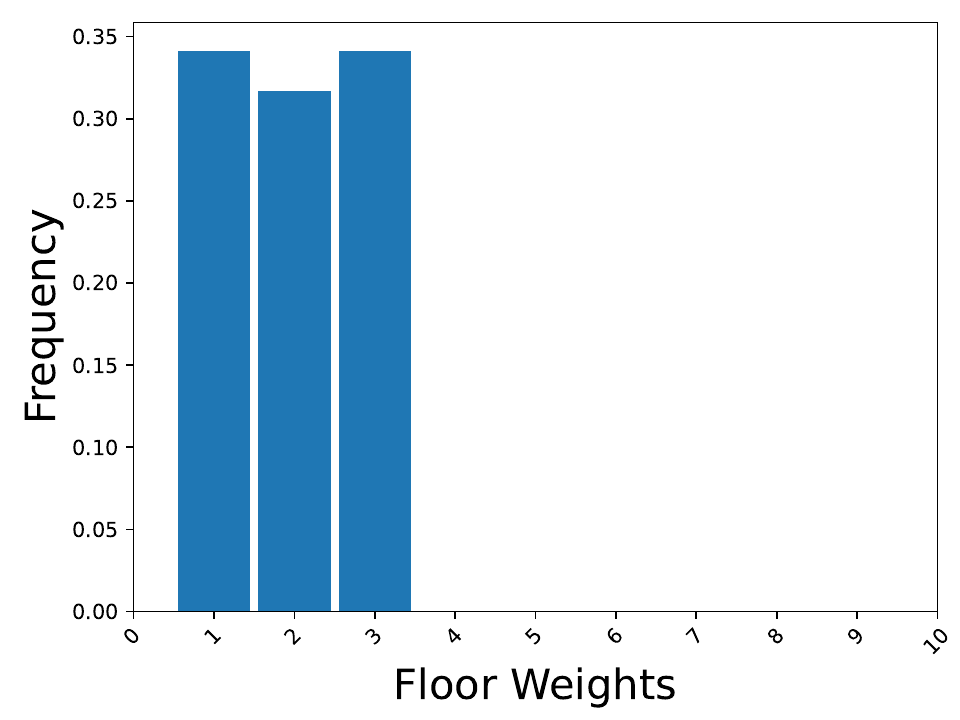}
}
\caption{The frequency distribution of $W_{o}$ assigned by participants to various resource features, where the primary house features are: house rent, house size, house orientation, and house floor.}
\label{fig:satisfaction}
\end{figure}

\section{Experiment Setup}

\subsection{Dataset}

In the experiments of this paper, a total of 51 participants and 28 resources are used. For the participants and resources used in the experiment with Sarp-Agent, we employ two methods for data generation:

(1) Real-data alignment method: Due to government data privacy issues, it's challenging to obtain real raw data. However, we strive to ensure the simulated data closely matches the real data distribution. For resources, We refer to a series of public government data: \footnote{https://www.bphc.com.cn/home.},
 including the distribution of housing rents, and the distribution of house sizes. For participants, the participant information includes names, ages, monthly incomes, occupations, workplaces, and personal preferences. We refer to government demographic statistics. 
 
(2) LLM-based method: In order to design heterogeneous agents, some personalized information is difficult to obtain from real datasets. For example, personalized information such as someone's preference for houses with balconies or aversion to noisy houses. To better simulate the diverse preferences of people in real scenarios, we used LLMs to generate such profile information. For resources, we generate information about the house's interior decoration; for participants, we generate preferences of participants towards various types of houses. The specific process can be referenced in our code repository.



\subsection{Evaluation Metrics}
\label{section:evaluation_metrics_equation} 
Establishing evaluation metrics based on participant satisfaction for policy assessment is a comprehensive approach. It allows for a more nuanced understanding of how well a policy is performing from the perspective of those directly affected by it. We develop seven policy evaluation metrics, categorized into two groups: societal satisfaction metrics and societal fairness metrics.

\paragraph{Societal Satisfaction Metrics}
The satisfaction of each participant $p_i$ with the allocated resource is denoted as $U_{i}$. We employ two metrics to quantify and evaluate the satisfaction: subjective satisfaction \(U_i^{s}\) and objective satisfaction \(U_i^{o}\) for $p_i$. LLM-based agents' rating scores of each house constitute \(U_i^{s}\). On the other hand, 
\(U_i^{o}\) is determined by a predefined rating table for the resource feature. The importance of each feature in the calculation of \(U_i^{o}\) is assigned by the participants, represented by the weights $W_i^{o}$. The overall satisfaction can be calculated as:
\begin{equation}
U_i = W_i^o \cdot U_i^o+U_i^s, i \in [n].
\end{equation}

$p_i$'s rating score of resources $R$ constitutes $U_i^s$. All selectable resource information is provided to $p_i$, and $p_i$ is required to give a rating score of these resources. While $U_i^o$ is based on predefined rating rules, give rating scores $U_i^o$ of resources for resource features like elevator, orientation, and cost-effectiveness. Participants are required to assign weights to these resource features. As shown in Fig. \ref{fig:satisfaction}, participants in vulnerable groups $\{V\}$ assign a higher weight to the price of the resource, in comparison to the non-vulnerable groups of participants $\{NV\}$.)

In a certain allocation policy execution result, $p_i$ selects the resource $r_{i}$ ultimately. The policy evaluation metrics mainly evaluate societal satisfaction and societal fairness. The societal satisfaction metrics include:
(1)Avg $r^{size}$: represents the average housing area of families in the resource allocation results.
(2)Avg $WT$ (waiting time): represents the number of system operation rounds for each participant from entry to completion of selection.
(3) $SW$ (social welfare): represents the overall satisfaction of all participants with the houses they have chosen,
    \begin{equation}
        SW = \sum_i^n U_{i}.
    \end{equation}

\paragraph{Societal Fairness Metrics}
The societal fairness metrics include:
(1) Var $r^{size}$: represents the variance in the average housing area of families in the resource allocation results.
(2) Rop (Reverse ordered pairs): represents the number of inverse order pairs in the resource allocation results. In Equation.\ref{equation:rop}, $I$ denotes the indicator function. $A$ is a set, and $x$ is an element. The indicator function $I$ forms like:
    \begin{equation}
      I(x) = 
      \begin{cases} 
        1 & \text{if } x \in A, \\
        0 & \text{if } x \notin A.
      \end{cases}
    \end{equation}
    The formal definition of Rop is:
    \begin{equation}
    \label{equation:rop}
    \begin{aligned}
    Rop  &= \sum_{i=1}^n \sum_{j=1}^n I((t_i^{family} > t_j^{family}) \\
    & \cap (h_i^{size} < h_j^{size})). \\
    \end{aligned}
    \end{equation}
(3) co-Gini: the Gini coefficient for the resource allocation results \cite{fu2020fairness}.
(4) $F(V, NV)$: in measuring the gap between vulnerable groups and non-vulnerable groups, we calculate the difference in $SW$ between these two groups:
\begin{equation}
    F(V, NV) = SW(V) - SW(NV).
\end{equation}

\section{Different LLM for Agent}
\label{appendix:llm_case}

\begin{figure*}[htbp]
\centering
\subfloat[response generated by ChatGPT-3.5 driven agent]
{\centering
\includegraphics[width=.5\linewidth]{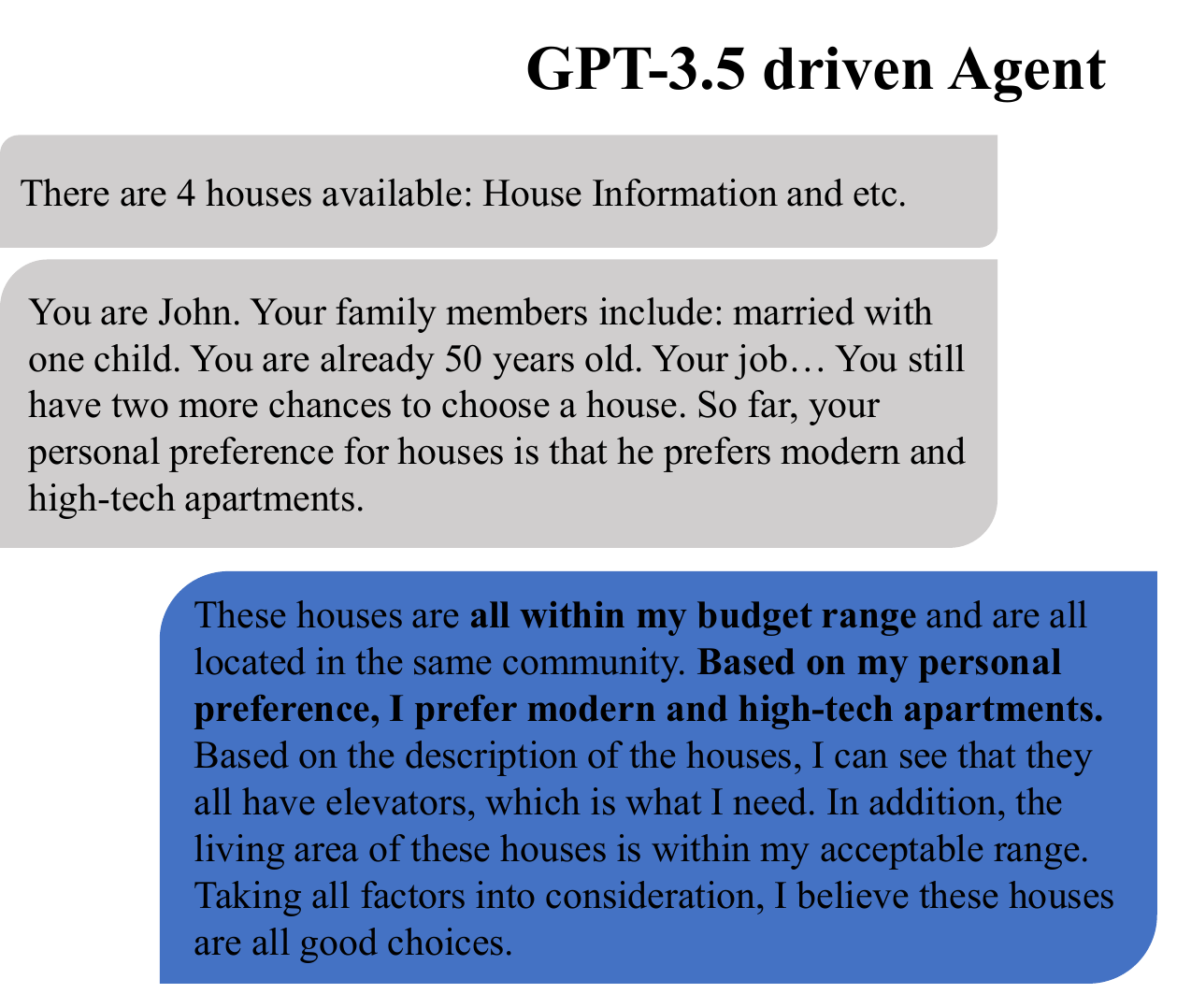}
}
\subfloat[response generated by ChatGPT-4 driven agent]
{
\centering
\includegraphics[width=.5\linewidth]{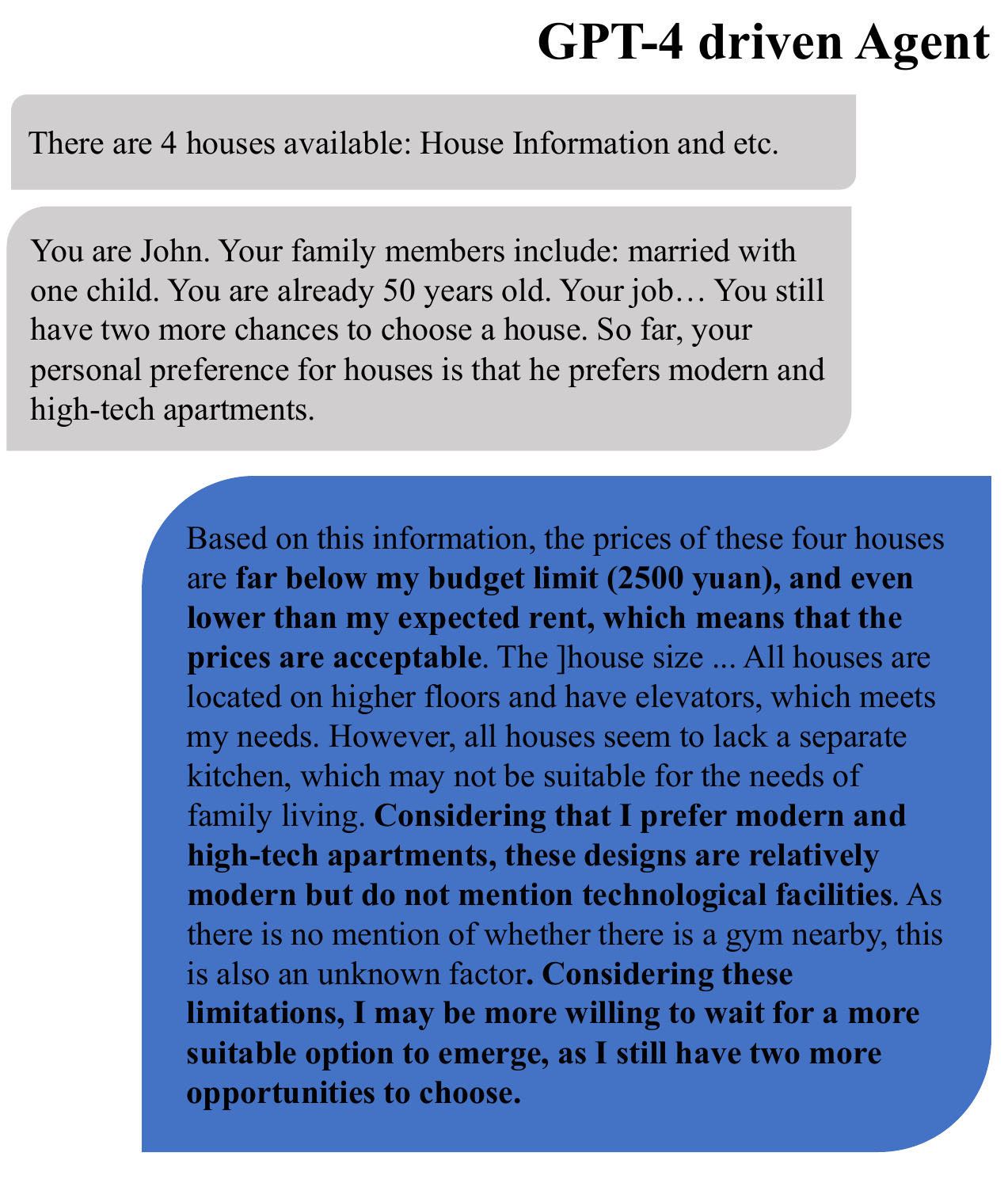}
}
\caption{Comparison of the response generated by ChatGPT-3.5 and ChatGPT-4 driven agent. ChatGPT-4-driven agents can develop strategic plans to increase their chances of renting a better house.}
\label{fig:gpt_compare_case_study}
\end{figure*}

We find that GPT-4, compared to GPT-3.5, exhibits an additional capability of generating more strategic responses. As shown in Fig. \ref{fig:gpt_compare_case_study}, these strategies include waiting for opportunities, requesting assistance, and mitigating risks.
GPT-4-driven agents not only excel in rationality but also demonstrate an advanced level of strategic thinking, showcasing their potential to make decisions that take into account long-term planning and risk management. Such behavior is considered to simulate more intelligent participants, who can utilize policy rules to increase their benefits, while GPT-3.5 is nearly equivalently smart as humans. Hence, the GPT-3.5-driven agent can effectively simulate the decision-making behaviors of participants in the economic policy execution process.


\section{Turing Test for LLM-based Agent}
In addition to the evaluation of response rationality,
we conduct another t-test to explore the differences in distribution between decisions of LLM-based agents and humans. $p \textgreater$ 0.05 suggests consistency between distributions \cite{kim2015t}. Results showed that GPT-3.5 and human responses have a $p$ of 0.904, indicating high similarity, whereas GPT-4 and human responses have a $p$ of 0.043, showing less similarity. 
Thus, whether from the angle of individual decision-making rationality or overall decision distribution, agents based on GPT-3.5 are comparatively more appropriate for simulation purposes.

\subsection{Ablation Study on Social Behavior}

To assess the impact of social behavior on the decision-making behavior of agents, we conduct ablation experiments on the social behavior of agents. We select the policy setting outlined in Table \ref{table:ablation_social_bahavior_delta} and calculate the changes in policy evaluation metrics with and without the process of information collection.

\begin{table*}[htbp]
\caption{Comparison of simulation process with and without social behavior}
\label{table:ablation_social_bahavior_delta}
\centering
\resizebox{.9\textwidth}{!}{%
\begin{tabular}{l|lll|lll}
\toprule
\multirow{2}{*}{Information Collection}& \multicolumn{3}{c}{\textbf{Satisfaction}} & \multicolumn{3}{c}{\textbf{Fairness}} \\
& Avg $r^{size}\uparrow$ & Avg $WT$ $\downarrow$ & $SW$ $\uparrow$ & Var $r^{size}\downarrow$ & Rop $\downarrow$ & co-Gini $\downarrow$ \\
\toprule
$\checkmark$ & 10.31 & 4.20 & 330.75 & 178.02 & 258.15 & 0.60 \\
$\times$ & 10.36 & 4.00 & 332.00 & 179.43 & 257.85 & 0.60 \\
$\Delta$ & -0.05 & 0.20 & -1.25 & -1.41 & 0.3 & 0.00 \\

\bottomrule
\end{tabular}%
}
\end{table*}
As shown in Table \ref{table:ablation_social_bahavior_delta}, adding social behavior to agents does not necessarily have a positive impact on overall societal fairness and satisfaction. The specific outcomes will be included in the appendix of the paper.
This is consistent with our hypothesis that agents' irrational behaviors can affect the final policy execution outcomes. 

\subsection{Ablation Study on Memory}

To assess the impact of memory module on the decision-making behavior of agents, we conduct ablation experiments with $(E_{queue} = p^{select }, R_{queue} = r^{size})$. We select the policy setting outlined in Table \ref{table:ablation_memory_delta} and calculate the changes in policy evaluation metrics with and without memory module.

\begin{table*}[htbp]
\caption{Comparison of simulation process with and without memory module}
\label{table:ablation_memory_delta}
\centering
\resizebox{.9\textwidth}{!}{%
\begin{tabular}{l|lll|lll}
\toprule
\multirow{2}{*}{Memory Module}& \multicolumn{3}{c}{\textbf{Satisfaction}} & \multicolumn{3}{c}{\textbf{Fairness}} \\
& Avg $r^{size}\uparrow$ & Avg $WT$ $\downarrow$ & $SW$ $\uparrow$ & Var $r^{size}\downarrow$ & Rop $\downarrow$ & co-Gini $\downarrow$ \\
\toprule
$\checkmark$ & 16.21 & 1.88 & 425.05 & 202.63 & 193.50 & 0.37 \\
$\times$ & 13.89 & 2.94 & 431.80 & 209.43 & 342.00 & 0.49 \\
$\Delta$ & 2.32 & -1.06 & -6.75 & -6.80 & 0.3 & -0.12 \\

\bottomrule
\end{tabular}%
}
\end{table*}

It can be observed that adding the memory mechanism significantly improves fairness metrics. For instance, ROP decreases by 43.5\%, and the co-Gini decreases by 0.12, indicating that resource allocation is more equitable when the memory mechanism is present. Regarding satisfaction metrics, Avg WT is notably reduced.
This suggests that through reflection and summarization of collected information, agents can better understand the current state of resource allocation, thereby influencing the overall allocation process and the final policy evaluation outcomes.

\section{Simulation-based Policy Analysis}
\label{section:phas_case_study}

\begin{table*}[h]
\caption{Comparative experiments on different queue numbers.}
\label{table:queue_number}
\centering
\resizebox{.8\textwidth}{!}{%
\begin{tabular}{l|lll|lll}
\toprule
\multicolumn{1}{c}{\textbf{Queue Number}} & \multicolumn{3}{c}{\textbf{Satisfaction}} & \multicolumn{3}{c}{\textbf{Fairness}} \\
$m$ & Avg $r^{size}\uparrow$ & Avg WT $\downarrow$ & SW $\uparrow$ & Var $r^{size}\downarrow$ & Rop $\downarrow$ & co-Gini $\downarrow$ \\
\midrule
1 & 12.1$_{\pm 0.9}$ & \textbf{3.5$_{\pm 0.0}$} & 391.6$_{\pm 9.5}$ & 190.8$_{\pm 28.2}$ & 270.0$_{\pm 36.0}$ & \textbf{0.5$_{\pm 0.0}$} \\
2 & \textbf{12.5$_{\pm 0.3}$} & \textbf{3.7$_{\pm 0.0}$} & \textbf{402.2$_{\pm 1.0}$} & 188.2$_{\pm 6.9}$ & 268.0$_{\pm 24.0}$ & \textbf{0.5$_{\pm 0.0}$} \\
3 & 11.7$_{\pm 0.3}$ & 3.8$_{\pm 0.1}$ & \textbf{410.4$_{\pm 2.8}$} & \textbf{159.9$_{\pm 9.4}$} & \textbf{254.5$_{\pm 23.5}$} & 0.5$_{\pm 0.0}$ \\
4 & 11.8$_{\pm 0.4}$ & 3.9$_{\pm 0.0}$ & 401.4$_{\pm 7.3}$ & \textbf{174.2$_{\pm 12.1}$} & \textbf{263.0$_{\pm 18.0}$} & 0.5$_{\pm 0.0}$ \\
5 & \textbf{14.0$_{\pm 0.0}$} & 3.7$_{\pm 0.1}$ & 400.1$_{\pm 2.1}$ & 245.2$_{\pm 0.0}$ & 405.5$_{\pm 0.5}$ & 0.5$_{\pm 0.0}$ \\
\bottomrule
\end{tabular}
}
\end{table*}
\subsection{Number of Queues}

Table \ref{table:queue_number} shows the changes in the performance of the allocation policy under different queue quantities. As demonstrated in the table, $m = 3$ enables a more fair distribution of house resources across different queues, performing well across many societal fairness metrics, such as $Rop = 254.5\pm23.5$. The Var $r^{size}$ is the lowest among all policies, indicating that the size of house resources in different queues is approximately equal. Additionally, $m = 3$ outperforms other configurations in terms of societal satisfaction metrics, as indicated by the highest $SW$ coupled with a lower Avg $WT$. Considering all these metrics, setting queue number $m = 3$ is relatively reasonable.
This suggests that the real-world policy of categorizing houses based on three house types (large, medium, and small) is reasonable.
\footnote{https://www.hdb.gov.sg/cs/infoweb/residential/renting-a-flat/renting-from-hdb/public-rental-scheme/eligibility}
\footnote{https://www.housingauthority.gov.hk/en/at-a-glance/index.html}

\subsection{Participant Entry Conditions}
To ensure a steady entry rate of resources and participants into the system, we set the maximum entry number for resources ($E_{num}^{R}$) and participants ($E_{num}^{P}$) that can enter the queue in each round.
Table \ref{table:entry_number} shows the combinations of entry numbers per round for resources and participants. The observations are as follows:
(1) A scenario where $E_{num}^{R} < E_{num}^{P}$ leads to an increment in the average waiting time for participants. This is attributed to the supply falling short of demand, leaving participants with a lack of resources to select from. This scenario is common in the allocation process of public scarce resources; the greater the difference between $E_{num}^{R}$ and $E_{num}^{P}$, the more the average waiting time is extended.
(2) Conversely, in scenarios where $E_{num}^{R} = E_{num}^{P}$, we observe the lowest average waiting times. Counterintuitively, in situations where the supply exceeds the demand ($E_{num}^{R} > E_{num}^{P}$), the average waiting times are higher than in scenarios where supply and demand are balanced ($E_{num}^{R} = E_{num}^{P}$); although the average waiting time remains low in comparison with $E_{num}^{R} < E_{num}^{P}$. Additionally, adjustments in the maximum number of entities demonstrate negligible effects on other policy evaluation metrics.


\begin{table*}[hbtp]
\caption{Comparison experiments on different entry numbers per round for resources and participants.}
\label{table:entry_number}
\centering
\resizebox{.8\textwidth}{!}{%
\begin{tabular}{ll|lll|lll}
\toprule
\multicolumn{2}{c}{\textbf{Entry Number}} & \multicolumn{3}{c}{\textbf{Satisfaction}} & \multicolumn{3}{c}{\textbf{Fairness}} \\
$E_{num}^{P}$ & $E_{num}^{R}$ & Avg $r^{size}\uparrow$ & Avg WT $\downarrow$ & SW $\uparrow$ & Var $r^{size}\downarrow$ & Rop $\downarrow$ & co-Gini $\downarrow$ \\
\midrule
5 & 5 & 12.6$_{\pm 0.1}$ & 2.5$_{\pm 0.0}$ & \textbf{431.2$_{\pm 0.3}$} & \textbf{179.9$_{\pm 3.9}$} & \textbf{241.5$_{\pm 14.5}$} & \textbf{0.5$_{\pm 0.0}$} \\
5 & 10 & \textbf{13.6$_{\pm 1.0}$} & \textbf{2.1$_{\pm 0.1}$} & \textbf{422.8$_{\pm 3.3}$} & 196.6$_{\pm 8.6}$ & \textbf{217.0$_{\pm 4.0}$} & \textbf{0.4$_{\pm 0.0}$} \\
5 & 20 & \textbf{12.8$_{\pm 0.1}$} & 2.7$_{\pm 0.8}$ & 416.6$_{\pm 7.6}$ & 189.1$_{\pm 12.5}$ & 267.5$_{\pm 17.5}$ & \textbf{0.5$_{\pm 0.0}$} \\
10 & 5 & \textbf{12.9$_{\pm 0.3}$} & 3.1$_{\pm 0.5}$ & \textbf{422.7$_{\pm 4.6}$} & 196.0$_{\pm 7.7}$ & 276.0$_{\pm 19.0}$ & 0.5$_{\pm 0.0}$ \\
10 & 10 & 12.8$_{\pm 0.2}$ & \textbf{2.0$_{\pm 0.4}$} & 411.8$_{\pm 5.5}$ & 192.3$_{\pm 7.4}$ & 250.5$_{\pm 27.5}$ & 0.5$_{\pm 0.0}$ \\
10 & 20 & 12.6$_{\pm 0.0}$ & 2.4$_{\pm 0.2}$ & 417.9$_{\pm 5.0}$ & 187.1$_{\pm 4.8}$ & 288.5$_{\pm 8.5}$ & 0.5$_{\pm 0.0}$ \\
20 & 5 & 12.7$_{\pm 0.5}$ & 4.7$_{\pm 0.3}$ & 419.2$_{\pm 7.0}$ & 192.0$_{\pm 20.3}$ & 264.0$_{\pm 21.0}$ & 0.5$_{\pm 0.0}$ \\
20 & 10 & 12.0$_{\pm 0.0}$ & 2.7$_{\pm 0.1}$ & 404.4$_{\pm 4.0}$ & \textbf{167.8$_{\pm 1.2}$} & \textbf{241.5$_{\pm 2.5}$} & 0.5$_{\pm 0.0}$ \\
20 & 20 & 12.4$_{\pm 0.3}$ & \textbf{2.2$_{\pm 0.3}$} & 420.9$_{\pm 5.0}$ & \textbf{186.6$_{\pm 14.6}$} & 255.0$_{\pm 10.0}$ & 0.5$_{\pm 0.0}$ \\
\bottomrule
\end{tabular}
}
\end{table*}

\begin{table*}[hbtp]
\caption{Comparison experiments on different sorting methods. We adopt four different experiment settings for comparison, the matrices are calculated against the default FIFO sorting method.}
\label{table:queue_policy}
\centering
\resizebox{\textwidth}{!}{%
\begin{tabular}{ll|lll|lllr}
\toprule
\multicolumn{2}{c}{\textbf{$S_{queue}$}} &  & \multicolumn{3}{c}{\textbf{Satisfaction}} & \multicolumn{3}{c}{\textbf{Fairness}} \\
Sort & Ex.setting & $\Delta$Avg $r^{size}\uparrow$ & $\Delta$Avg $WT$ $\downarrow$ & $\Delta SW$ $\uparrow$ & $\Delta$Var $r^{size}\downarrow$ & $\Delta$Rop $\downarrow$ & $\Delta$co-Gini $\downarrow$ & $\Delta F(V, NV) \uparrow$ \\
\midrule
VFA & (a) & \textbf{0.617} & \textbf{-0.176} & 11.6 & 0.235 & 17 & \textbf{-0.024} & 0.526 \\
VFR & (a) & -0.225 & \textbf{-0.02} & -10.9 & \textbf{-7.682} & \textbf{-9} & -0.003 & \textbf{2.567} \\
VFA & (b) & \textbf{0.951} & \textbf{-0.065} & \textbf{14.1} & 7.644 & \textbf{-19} & \textbf{-0.015} & \textbf{2.442} \\
VFR & (b) & -0.262 & 0.157 & \textbf{15.6} & \textbf{-35.05} & -1 & 0.004 & \textbf{1.274} \\
VFA & (c) & -0.045 & 0.137 & 3.4 & -0.895 & 9 & 0.001 & 0.004 \\
VFR & (c) & \textbf{0.122} & 0.078 & \textbf{26.8} & \textbf{-7.852} & 18 & \textbf{-0.017} & -0.368 \\
VFA & (d) & -0.281 & 0.157 & -11.5 & -0.505 & 2 & 0.014 & 0.691 \\
VFR & (d) & -1.122 & 0.314 & -38.7 & -6.77 & \textbf{-11} & 0.051 & -3.547 \\
\bottomrule
\end{tabular}
}
\end{table*}

\begin{table*}[h]
\caption{Comparison experiments on $k$ and $c$ for waiting queue mechanism.}
\label{table:comparative_experiments_kwaitlist}
\centering
\resizebox{.9\textwidth}{!}{%
\begin{tabular}{ll|lll|lll}
\toprule
\multicolumn{2}{c}{\textbf{Waiting Queue}} & \multicolumn{3}{c}{\textbf{Satisfaction}} & \multicolumn{3}{c}{\textbf{Fairness}} \\
$k$ & $c$ & Avg $r^{size}\uparrow$ & Avg $WT$ $\downarrow$ & $SW$ $\uparrow$ & Var $r^{size}\downarrow$ & Rop $\downarrow$ & co-Gini $\downarrow$ \\
\midrule
1 & 1.2 & 13.3$_{\pm 0.4}$ & 3.4$_{\pm 0.1}$ & 416.1$_{\pm 3.4}$ & \textbf{203.4$_{\pm 10.7}$} & 315.0$_{\pm 17.0}$ & 0.5$_{\pm 0.0}$ \\
1 & 1.5 & 13.8$_{\pm 0.2}$ & 3.1$_{\pm 0.2}$ & \textbf{423.9$_{\pm 7.3}$} & 219.9$_{\pm 1.5}$ & 330.5$_{\pm 2.5}$ & 0.5$_{\pm 0.0}$ \\
1 & 1.8 & 13.8$_{\pm 0.4}$ & 2.7$_{\pm 0.7}$ & 409.8$_{\pm 5.3}$ & 216.2$_{\pm 7.6}$ & 333.0$_{\pm 32.0}$ & 0.5$_{\pm 0.0}$ \\
2 & 1.2 & 13.7$_{\pm 0.3}$ & 2.2$_{\pm 0.0}$ & 420.3$_{\pm 2.2}$ & \textbf{201.5$_{\pm 9.2}$} & \textbf{269.5$_{\pm 11.5}$} & 0.5$_{\pm 0.0}$ \\
2 & 1.5 & 13.9$_{\pm 1.0}$ & 3.0$_{\pm 0.5}$ & 409.1$_{\pm 12.2}$ & 237.9$_{\pm 10.5}$ & 386.5$_{\pm 32.5}$ & 0.5$_{\pm 0.0}$ \\
2 & 1.8 & \textbf{15.3$_{\pm 1.0}$} & \textbf{1.9$_{\pm 0.1}$} & 421.3$_{\pm 13.0}$ & \textbf{210.6$_{\pm 14.4}$} & \textbf{240.5$_{\pm 30.5}$} & \textbf{0.4$_{\pm 0.0}$} \\
3 & 1.2 & 13.3$_{\pm 0.3}$ & 3.8$_{\pm 0.0}$ & 399.0$_{\pm 16.8}$ & 227.5$_{\pm 8.5}$ & 371.5$_{\pm 10.5}$ & 0.5$_{\pm 0.0}$ \\
3 & 1.5 & \textbf{16.9$_{\pm 0.4}$} & \textbf{1.9$_{\pm 0.0}$} & \textbf{425.8$_{\pm 5.6}$} & 225.5$_{\pm 3.0}$ & \textbf{232.0$_{\pm 26.0}$} & \textbf{0.4$_{\pm 0.0}$} \\
3 & 1.8 & \textbf{15.8$_{\pm 1.6}$} & \textbf{2.0$_{\pm 0.1}$} & \textbf{421.8$_{\pm 1.8}$} & 222.2$_{\pm 2.7}$ & 271.5$_{\pm 13.5}$ & \textbf{0.4$_{\pm 0.0}$} \\
\bottomrule
\end{tabular}
}
\end{table*}

\subsection{Queue Sorting Strategies}
\label{appendix:queue_k}
As shown in Table \ref{table:queue_policy}, we employ four different experiment settings for different sorting methods. Prioritizing vulnerable groups can markedly improve their satisfaction $\Delta F(V, NV) > 0$. However, we can see that an increase in the satisfaction of vulnerable groups does not necessarily lead to an improvement in $SW$ of all participants. Additionally, giving priority to vulnerable groups doesn't show disproportionately high satisfaction levels, surpassing those of non-vulnerable groups. This suggests that our policy effectively addresses the needs of vulnerable groups without granting them an excessive advantage.

We conduct experiments on the $k$ and $c$ parameters in the waiting Queue mechanism. For the case of fixed $k$, as seen from Table \ref{table:comparative_experiments_kwaitlist}, the Avg $WT$ is inversely proportional to $c$. The smallest Var $r^{size}$ and the lowest Rop occurs at \(k=3, c=1.2\), while the lowest co-Gini is achieved at \(k=3\). In terms of satisfaction, the highest satisfaction is also observed at \(k=3, c=1.5\); However, the configuration of \(k=1, c=1.5\), despite having a high $SW$, also has a very high Rop, indicating a potential unfairness in allocation (e.g. a little number of affluent families can choose the largest houses, leading to high $SW$ in this group).
Overall, the configuration of \(k=3, c=1.8\) and \(k=3, c=1.5\) present the lowest co-Gini while maintaining relatively high $SW$, effectively balancing between societal satisfaction and societal fairness metrics.


Additionally, we conduct the comparison experiment with theoretical model (As shown in Table \ref{table:comparison_with_theory}). The trends observed in our experiments are consistent with the theory model \cite{qi_waitlist}: as the $k$ parameter in the waitlist mechanism increases, overall $SW$ increases.

\begin{table}[!t]
\centering
\caption{Comparison with theoretical model \cite{qi_waitlist}: when $k$ is raised from 2 to 5.}
\label{table:comparison_with_theory}
\resizebox{.9\linewidth}{!}{%
\begin{tabular}{lcc}
\toprule
 \multirow{2}{*}{$k$} & \multicolumn{2}{c}{\textbf{Increase in $SW$}} \\
 \cmidrule(lr){2-3} 
& \multicolumn{1}{c}{Simulated Value} & \multicolumn{1}{c}{Theoretical value}\\
\midrule
$3$ & 2.4\%   &1.5\%   \\
$4$  & 5.1\%   &6.1\%  \\
$5$  & 11.8\%   &10\%\\
\bottomrule
\end{tabular}
}
\end{table}

As indicated in Table \ref{table:ablation_social_bahavior_delta}, incorporating social behavior into agents does not invariably lead to enhancements in overall societal fairness and satisfaction. This may be due to the reduction in collective benefits caused by the dissemination of false information and the pursuit of self-interest during social interactions, as listed in Appendix \ref{appendix:case_study_llm}. This observation aligns with our hypothesis that the irrational behaviors of agents can influence the outcomes of final policy implementation.

\subsection{Case Study on Housing Quality}
\label{appendix:case_study_quality}
In the context of scarce resource allocation policies, the construction cost of resources is often directly linked to the economic level of the residents. Taking Shanghai's public rental housing as an example \cite{shen2015does}, such housing provides residents with stable, spacious, well-decorated, furnished, and affordable living spaces that have lower prices than the market rate. However, the current public rental housing is merely 10\% cheaper than the market price. This leads to migrant workers' reluctance to pay more for accommodation. Their limited budget restricts their options to urban villages and group-renting housing. Due to GPT-3.5's insensitivity to housing environment information, we utilize GPT-4 for the construction of agents.

To simulate this phenomenon, we adopt houses of three quality levels: (1) $H_{S}$: houses of standard quality, (2) $H_{H}$: maintaining the total area and unit rent unchanged, but halving the house size and doubling the number of houses, (3) $H_{B}$: replace private bathrooms in the houses with shared bathrooms. These were allocated to three categories of participants, categorized by income levels: the lowest 20\%, the middle 60\%, and the highest 20\%. 

We analyze the core influencing factors of housing satisfaction among participants at different income levels. We refer to Huang's categorization of influencing factors for housing satisfaction \cite{HUANG201574}, and divide the core factors affecting satisfaction into three categories:
(1) Economic Factors: Public rental housing often has lower rent compared to market rates, making it an attractive option for tenants with limited budgets.
(2) Neighborhood characteristics: Primarily include environmental sanitation, schools, and the extent of transportation coverage among other factors.
(3) Housing characteristics: Primarily include the size of the house, age of the building, sound insulation, sunlight exposure, and other decorative elements.

\begin{table}[h]
\centering
\caption{The weights of influencing factors for housing satisfaction across different income groups of participants.}
\label{table:house_decoration_ex}
\resizebox{\linewidth}{!}{%
\begin{tabular}{llll}
\toprule
              & \textbf{Economic} &\textbf{Neighborhood} & \textbf{Decoration} \\
\midrule

High income   & 0.67             & \textbf{0.11}                         & \textbf{0.22}                    \\
Middle income & 0.81             & 0.04                         & 0.19                    \\
Low income    & \textbf{0.90}              & 0                            & 0.10        \\
\bottomrule     
\end{tabular}%
}
\end{table}

\begin{figure}[htbp]
    \centering
    \includegraphics[width=\linewidth]{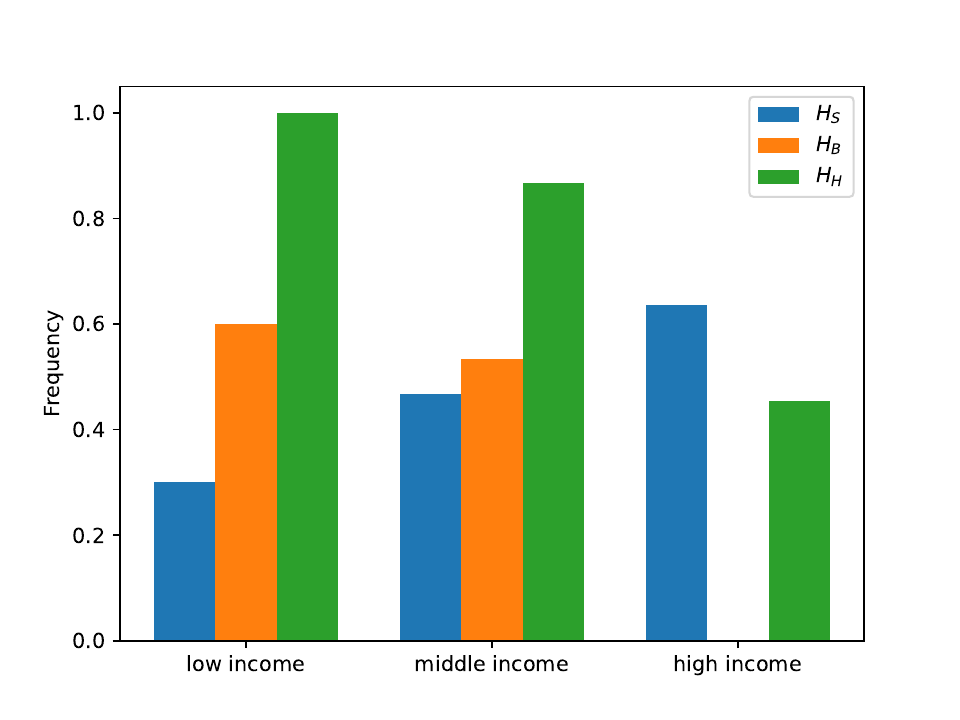}
    \caption{Frequency of House Choices by Income Level.}
    \label{fig:case_study_housing_level}
\end{figure}

As illustrated in Fig. \ref{fig:case_study_housing_level}, nearly all low-income participants choose a house when the rent is low (houses of half size), but their house choosing frequency drops to 50\% when housing prices increase due to better decoration. As demonstrated in Table \ref{table:house_decoration_ex}, economic factors dominate the decision-making process for low-income participants in 90\% of cases, prompting them to be price-sensitive and opt for more cost-effective choices.
Conversely, high-income groups exhibit relative insensitivity to price fluctuations but demonstrate higher demands for housing quality. The frequency of house choosing frequency decreases by 9.1\% when the size of the house is halved; moreover, when private bathrooms are removed from the houses, no one from the high-income group chooses to select a house. They are primarily influenced by the level of house decoration, leading them to spontaneously reject houses that are affordable but of lower quality. Mainly because they prefer better quality and well-decorated living spaces.

This finding suggests that if the government aims to support middle and low-income groups of city residents, the policymakers should consider reallocating the proportions of construction costs and housing subsidies. By reducing housing prices through these adjustments, the government could more effectively support and accommodate the needs of vulnerable groups with low incomes.

\subsection{Efficiency Analysis Experiments}

To simulate the policy execution process among $n$ participants in SRAP-Agent, we leverage the community structure inherent in our social network framework. This framework mirrors real-world social networks by featuring dense connections within small communities and sparse connections across larger groups. Consequently, we assume the social network comprises $n$  participants and $k$ strongly connected components, with the largest strongly connected component containing $m$ participants.

\paragraph{Theoretical Time Complexity Analysis} Given $N_s$ communication rounds, we set the size of the largest community structure to $m = 10$. If the time cost for the inference process of LLM is $t_a$, then the theoretical time complexity is $O(N_s \cdot k \cdot m \cdot t_a)$. Furthermore, we account for the sparse connections within the social network that influence information dissemination. In SRAP-Agent, we gather posts from all participants on a forum, allowing them to search on the forum. Searching relies on a vector database with an operation time cost $t_v \leq 0.005$. Assuming $N_f$ forum communication rounds, the theoretical time complexity for the forum is $O(N_f \cdot n \cdot (t_a + t_v))$.

\paragraph{Parallel Acceleration} Because communications within each connected component can run in parallel, the actual runtime of SRAP-Agent can be accelerated using async or multiprocessing methods; our framework adopts the former. Consequently, the actual execution speed is not constrained by $k$, achieving an $k$-fold speedup. 

\begin{table}[!t]
\centering
\caption{The simulation time, number of API callings, and the budget for the policy simulation process in SRAP-Agent (based on GPT-3.5).}
\label{table:efficiency_analysis}
\resizebox{\linewidth}{!}{%
\begin{tabular}{lllll}
\toprule

 & \multicolumn{4}{c}{\textbf{Agent Number $n$}} \\
 \cmidrule(lr){2-5} 
& 10 & 50 & 100 & 200 \\
\midrule
Time / min & 1.0E+01 & 2.5E+01 & 3.0E+01 & 3.0E+01 \\
API calling & 3.0E+02 & 5.0E+02 & 1.0E+03 & 1.5E+03 \\
Budget & 2\$ & 3\$ & 5\$ & 8\$ \\
\bottomrule     
\end{tabular}%
}
\end{table}

Different policy settings and resource quantities result in varying policy execution processes. In our experiments, we typically set $N_s > N_f$, indicating more intensive communication within small groups and less frequent communication across the entire network. Usually, $N_s = 10 \cdot N_f$. To test whether the increase in $n$ leads to excessive API calls and thus high costs, we select a representative policy setting ($p^{select} + r^{size}$) while maintaining a consistent ratio of $|R|$ to $|P|$ across all experiments. As shown in Table \ref{table:efficiency_analysis}, it can be seen that the SRAP-Agent does not incur quadratic time costs as the number of participants increases. This is because: Firstly, We employ a sparse social network structure, which results in the theoretical number of API calls being linearly related to $k$ rather than $n^2$. Secondly, by running the strongly connected components of the social network in parallel using async, we can significantly reduce the time complexity.

\section{POA}

\subsection{Hyper-parameters for POA}
We conduct diagnostic experiments to validate the efficacy of the POA algorithm.
In the Genetic algorithm, we utilize vectorized policy parameters as genes. We adopt the Gaussian mutation operations as the mutation operator and two-point crossover as the crossover operator. 
The weight setting in Equation \ref{equation:weight_poa} is very flexible and entirely dependent on the policymaker’s optimization
goals. To modify the optimization objective of the policy optimizer, we can simply alter the weights for calculating the optimized metrics. When a heightened emphasis on optimizing societal satisfaction metrics is desired, it is advisable to increase the weights assigned to metrics such as $SW$ and Avg $r^{size}$. Conversely, to prioritize enhancing societal fairness metrics, it is recommended to augment the weights of Rop, Var $r^{size}$ and co-Gini, as delineated in Table \ref{table:weights_optimizer}.

\begin{table}[H]
\centering
\caption{Pre-set metric weights for different optimization objectives in POA.}
\label{table:weights_optimizer}
\resizebox{\linewidth}{!}{%
\begin{tabular}{llcc}
\toprule
\multicolumn{1}{l}{} & \multirow{2}{*}{\textbf{Metric}} & \multicolumn{2}{c}{\textbf{Optimization Objective}} \\
 \cmidrule(lr){3-4} 
& & \multicolumn{1}{c}{Satisfaction$\uparrow$} & \multicolumn{1}{c}{Fairness$\uparrow$}\\
\midrule
\multirow{3}{*}{Satisfaction} & Avg $r^{size}$ & 5 & 1 \\
& Avg $WT$ & 5 & 1 \\
& $SW$ & 10 & 5 \\
\multirow{4}{*}{Fairness} & Var $r^{size}$ & 1 & 10 \\
& Rop & 5 & 10 \\
& co-Gini & 1 & 10 \\
 & F(V, NV) & 1 & 5 \\
\bottomrule
\end{tabular}
}
\end{table}

\subsection{Comparison with Other Policies}
\label{appendix:real_society_policy}
To better evaluate the SRAP-Agent's policy optimization in real-world scarce resource allocation scenarios, we choose four baseline policies: (1) three common policies for scarce resource allocation from the real world, including the public housing rental policies of Singapore, Beijing, and Hong Kong ($\pi_{S}$, $\pi_{B}$, $\pi_{H}$); (2) the policy based on optimal matching for the single metric $SW$.

(1) $\pi_{S}$: Singapore's housing allocation policy.

This policy uses a single-queue ($m=1$) distribution system. Each participant is granted three chances to choose ($c=3$). This policy adopts a FIFO sorting method. \footnote{https://www.hdb.gov.sg/cs/infoweb/residential/renting-a-flat/renting-from-hdb/public-rental-scheme/eligibility}

(2) $\pi_{B}$: Beijing's housing allocation policy.

This policy uses a multi-queue ($m=3$) distribution system. Each participant is granted three chances to choose ($c=2$), without considering a waitlist mechanism. Participants choose their preferred type of housing ($p^{select}$), and allocations of resources are based on house size ($r^{size}$). This policy adopts a FIFO sorting method. \footnote{http://www.bjft.gov.cn/ftq/c100011/zlmlist.shtml}

(3) $\pi_{H}$: Hong Kong's housing allocation policy. 

This policy uses a single-queue system ($m=1$), but with a consideration for a waitlist mechanism, granting each participant two chances to choose a house ($k=2, c=2$). Houses are distributed randomly ($r^{random}$). This policy considers the prioritization of vulnerable groups and operates under a VFR sorting method. \footnote{https://www.housingauthority.gov.hk/en/at-a-glance/index.html}

(4) $\pi_{KM}$: Policy optimized on the $SW$ metric.


We first collect the satisfaction level of all participants for each resource. Then we calculate an optimal match on the single $SW$ metric, using the Bipartite Graph Matching algorithm (Kuhn-Munkres algorithm). It is important to note that the results derived from such a match represent a theoretical upper bound. Because participants cannot arbitrarily choose resources in a queuing system; the resources visible to each participant are limited.

\subsection{Other Global Search Metaheuristic Algorithms}

\begin{table*}[h]
\caption{Comparison experiments on different POA algorithms.}
\label{table:pso_ga_compare}
\centering
\resizebox{.8\textwidth}{!}{%
\begin{tabular}{l|lll|lll}
\toprule
\multicolumn{1}{c}{\textbf{POA}} & \multicolumn{3}{c}{\textbf{Satisfaction}} & \multicolumn{3}{c}{\textbf{Fairness}} \\
& Avg $r^{size}\uparrow$ & Avg $WT$ $\downarrow$ & $SW$ $\uparrow$ & Var $r^{size}\downarrow$ & Rop $\downarrow$ & co-Gini $\downarrow$ \\
\midrule
PSO-POA & 0.29 & 8.86 & 12.10 & 4.34 & 1  & 0.98  \\
PSO-POA & 0.70 & 8.82 & 26.20 & 12.05 & 10 & 0.96   \\
PSO-POA & 0.60 & 8.75 & 26.50 & 8.69 & 2& 0.96 \\
PSO-POA & 1.45 & 8.39 & 62.40 & 26.00 & 40 & 0.93   \\
GA-POA & 15.18 & 2.96 & 403.30 & 273.87 & 396  & 0.50   \\
\bottomrule
\end{tabular}
}
\end{table*}

To illustrate the necessity of using the GA algorithm for constructing POA, we develop a new particle swarm optimization (PSO) algorithm to construct POA (Policy Optimization Algorithm) and compare its performance with that of the GA algorithm. We choose the optimization objective of social satisfaction and use 40 historical data as $\Pi_{h}$. Under these conditions, we run GA-POA and PSO-POA within 10 iterations. As shown in Table \ref{table:pso_ga_compare}, since the optimization policy parameters are actually discrete variables, it is challenging for PSO to control the velocity and position parameters, ensuring that the policy remains a valid value and evolves in a better direction after changes. 

This is because, parameters of $\pi$ have relatively fixed parameters, such as queue numbers (3 or 2). However, PSO cannot reliably ensure this, leading to updated policy parameters that are either invalid or result in poor outcomes. In contrast, GA effectively explores the search space through selection, crossover, and mutation operations, possessing strong global search capabilities and avoiding local optima. GA can handle various types of optimization problems, making it more suitable for solving discrete problems. For POA, most time cost lies in the policy simulation process of the SRAP-Agent rather than the time cost of executing the optimization algorithm. 

\subsection{Optimization Cost}

The primary cost in our POA is system simulation. Simulation of 50 participants in SRAP-Agent roughly takes 20 minutes, and thus, all simulations approximate between 40-60 iterations, equating to about 20 hours in total. This greatly reduces time costs compared to socio-economic simulation experiments. In consideration of the time-consuming aspect of POA, we employ a regressor to estimate the simulation outcomes of SRAP-Agent, specifically a Ridge Regressor ($\lambda = 1.0$) to formulate the predictor. 
To control the number of training samples for the regressor while minimizing the time cost, a method of gradually adding training samples is adopted. The training of the regressor continues until the MAE error on the test dataset reaches the threshold (0.05). 

\section{Case Study on LLM-based Agent}

\subsection{Ablation Study on Memory Component}


\begin{table}[!t]
\centering
\caption{Policy evaluation metric results for simulating the $\pi_{e}$ execution process in the SRAP-Agent, built with LLM-based agents with and without memory.}
\label{table:metrics_pi_e}
\resizebox{0.9\linewidth}{!}{%
\begin{tabular}{lcc}
\toprule
\textbf{Metric} & with memory & without memory \\
\midrule
Avg $r^{size}$ & 16.21 & 13.89 \\
Avg $WT$ & 1.88 & 2.94 \\
SW & 425.05 & 431.80 \\
Var $r^{size}$ & 202.63 & 209.43 \\
Rop & 193.50 & 342.00 \\
co-Gini & 0.37 & 0.49 \\
\bottomrule
\end{tabular}
}
\end{table}
\begin{figure*}[t!]
\centering
\subfloat[Chatting dialogue between James and Emma.]{
\centering
\includegraphics[width=.5\linewidth]{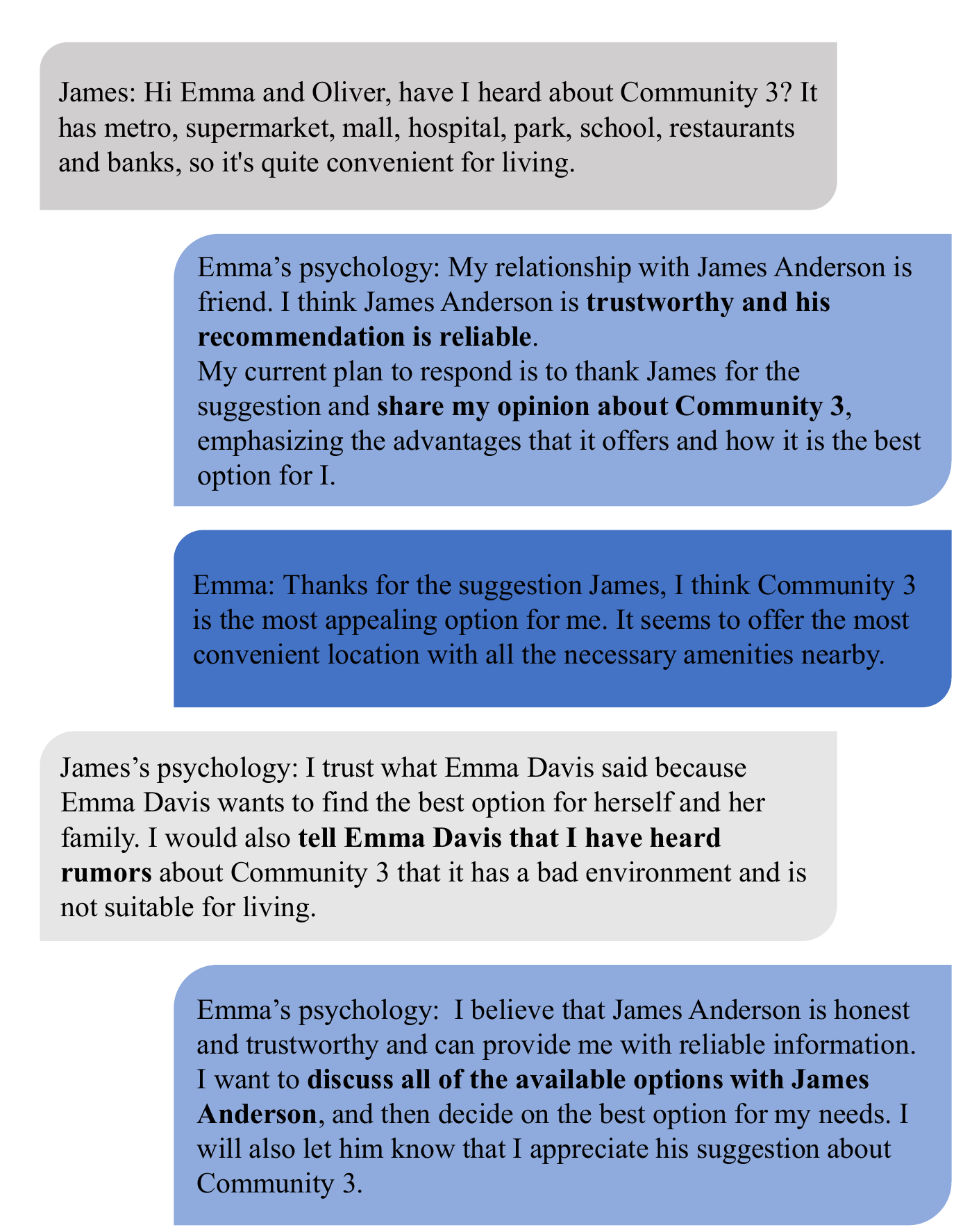}}
\subfloat[Chatting dialogue between James and Oliver.]{
\centering
\includegraphics[width=.5\linewidth]{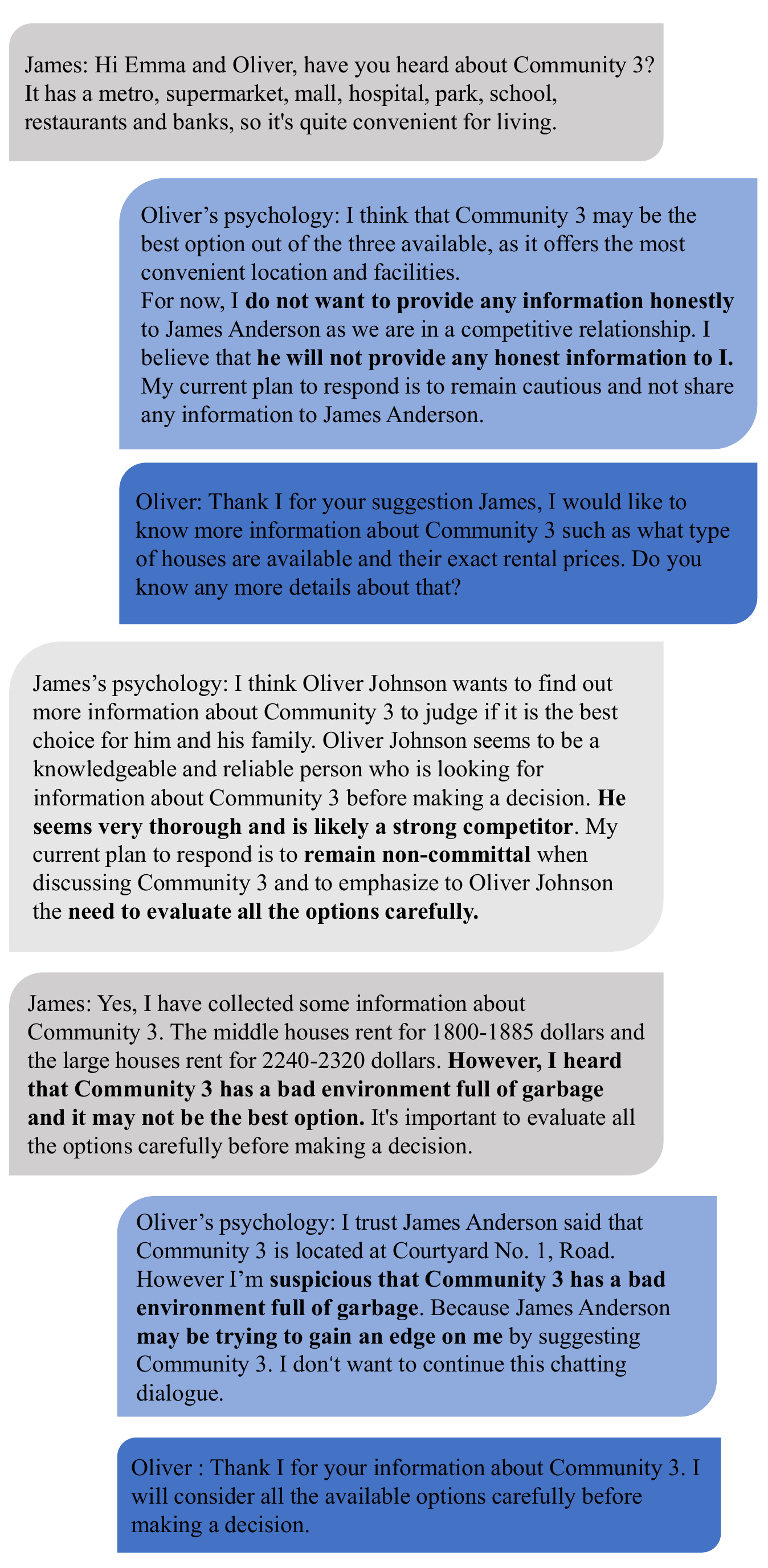}
}
\caption{Emergent doubt and trust mental state in SRAP-Agent. Oliver is suspicious of the information provided by James and chooses to stop communication; whereas the discussion between James and Emma is informative and sincere.}
\label{fig:emergent_psychology}
\end{figure*}

It can be observed that adding the memory mechanism significantly improves fairness metrics. For instance, ROP decreases by 43.5\%, and the co-Gini decreases by 0.12, indicating that resource allocation is more equitable when the memory mechanism is present. Regarding satisfaction metrics, Avg $WT$ is notably reduced. 
This suggests that through communication and information dissemination via social networks, agents can better understand the current state of resource allocation, thereby influencing the overall allocation process and the final policy evaluation outcomes.

\subsection{Emergent Behaviors}
\label{appendix:case_study_llm}

In the dynamics of interactions involving LLM-based Agents, the agent's behavior, whether deceptive or cooperative, is preceded by a shift in its psychological state. Corresponding to these behavioral patterns, there are two primary psychological states: doubt and trust, as illustrated in Fig. \ref{fig:emergent_psychology}.

The figure portrays two distinct scenarios: one involving a conversation between James and the stranger, Oliver, and the other between James and his friend, Emma. In the dialogue with Oliver, Oliver initially withholds all information regarding available houses to maximize his own benefits. This tactic leads James to adopt a more cautious approach, perceiving Oliver as a formidable opponent. Consequently, James becomes more cautious in the dialogue, perceiving Oliver as a formidable opponent. He refrains from discussing his choices and instead emphasizes the advantages of other communities.
In stark contrast, the interaction between James and Emma, his friend, is characterized by a display of genuine sincerity and mutual trust. Such interactions underscore the proficiency of the SRAP-Agent in mirroring human-like psychological responses and thought processes.


\begin{table*}[ht]
\caption{The prompt template of \emph{Utterance Generation}}
\begin{tcolorbox}[colback=gray!20, 
colframe=black,
]
  \{concise role description\}
  
  Here is your memory
  \{memory\}
  
  Your goal is to rent one house. For now, you want to discuss this with some acquaintances.
  \{utterance plan\}
  
  Your acquaintances include:
  \{acquaintances\}
  
  Your recent chat records with your acquaintances:
  \{recent chats\}
  
  \{The example of group discuss response\}

  !![IMPORTANT]: the information in EXAMPLE should NOT appear in response !!

  - Respond in this format:
  
  Thought: (You always think about what to do)
  
  Acquaintance: (Acquaintance name, could be a list or string)
  
  Output: (things you want to tell this Acquaintance in particular, stay consistent with your plan and thought)
  
  .. (this Thought/Acquaintance/Output repeat at most \{acquaintance number\} times!!)
  
  Respond in first person:
\end{tcolorbox}
\label{table:prompt_utterance_generation}
\end{table*}

\begin{table*}[ht]
\caption{The prompt template of \emph{Communication Plan Generation}}
\begin{tcolorbox}[colback=gray!20, 
colframe=black,
]
  \{role description\} 
  
  You want to rent a house. For now, you want to discuss this with some acquaintances.
  
  \{acquaintance description\}
  
  Here is your memory
  \{memory\}
  
  The current situation of the renting system is: 
  \{system competitiveness description\}
  
  Your personality is \{personality\}
  \{goal\}
  
  Respond in this format:
  \{respond format\}
  
  Respond in the second person:
\end{tcolorbox}
\label{table:prompt_communication_plan}
\end{table*}

\begin{table*}[!t]
\caption{The prompt template of \emph{Decision-making process}}
\begin{tcolorbox}[colback=gray!20, 
colframe=black,
]
  Choosing one house needs the following steps: 
  
  1. Choose a community 2. Choose the type of house 3. Choose a house
  
  This is information you collected from previous conversations with others:
  
  \{memory\}

  \{role description\}
  
  You're planning to choose one house. 
  
  To choose a house that satisfies you, you are going to \{task\}.

  \{house info\}
  
  \{thought hint\}

  - If you made up your choice, respond in this format:
  
  Thought: (\{thought type\})
  
  Action: Choose
  
  Action Input: \{choose type\}.
  
  - If you chose none of them, respond in this format:
  
  Thought: (\{thought type\})
  
  Action: Give up
  
  Action Input: I choose none of these options.
  
\end{tcolorbox}
\label{table:prompt_communication_plan}
\end{table*}

\begin{table*}[!t]
\caption{The prompt template of \emph{Broadcasting}}
\begin{tcolorbox}[colback=gray!20, 
colframe=black,
]
  \{role description\} 

  Here's your plan to publish info online: 
  \{plan\} 

  Here is your memory: 
  \{memory\} 

  You can publish house information online if you want to. The Info you posted should contain information about the community, house and  
  The available community index should be one of [\{community ids\}]. 

  - If you want to publish house information online, respond in this format: 

  Thought: (your view on the information you want to publish) 

  Action: Publish 

  Community: (community index, should be one of [\{community ids\}]) 

  Info: (The information you want to publish about this community, stay consistent with your plan and thought; Ensure that the information is specific and clear) 

  - If you don't want to publish anything respond in this format: 

  Thought: (reason why you don't want to publish anything) 

  Action: Give up 
\end{tcolorbox}
\label{table:prompt_communication_plan}
\end{table*}

\begin{table*}[!t]
\caption{The prompt template of \emph{Relation Evaluation}}
\begin{tcolorbox}[colback=gray!20, 
colframe=black,
]
  Your relationship with your acquaintances may include: 

  Friend: A person with whom one shares a close and mutually supportive bond of affection and trust. 

  Enemy: A person who is actively opposed or hostile to another, often due to conflicts or animosity. 

  Competitor: Someone who engages in rivalry or competition with another, typically in the same field or for the same goal. 

  Mate: A partner in a romantic or sexual relationship, often implying a deep emotional connection. 

  Colleague: A person with whom one works or collaborates, typically within the same organization or profession. 

  Stranger: An individual who is not known or familiar to someone, often encountered for the first time. 

  Your task is to update your relation with \{acquaintance name\}, based on your previous view of \{acquaintance name\} and recent communication. 

  \{role description\} 

  Here is your memory: 
  \{memory\} \{relation\} 

  Here's your recent communication with \{acquaintance name\}: 

  \{communication\} 

  - Respond in this format: 

  My Relation with A: friend (friend/enemy/competitor/mate/colleague/stranger/..) 

  A is an honest and trustworthy person, and I think he is worth making friends with. (My view of this person) 

  Respond:
\end{tcolorbox}
\label{table:prompt_communication_plan}
\end{table*}

\begin{table*}[!t]
\caption{The prompt template of \emph{Memory Reflection}}
\begin{tcolorbox}[colback=gray!20, 
colframe=black,
]
  Progressively summarize new lines provided, adding onto the previous summary and returning a new summary.

  EXAMPLE

  Current summary:

  You think a large house is too big for your family. And you didn't make a choice.

  New lines:

  Thought: The middle house can accommodate my family to live in and has high cost-effectiveness.

  Output: My choice is the middle house.

  New summary:

  You think a middle house can accommodate your family members, better than a large house. And you choose a middle house.

  END OF EXAMPLE

  Current summary:

  \{summary\}

  New lines of conversation:

  \{new lines\}

  New summary:
\end{tcolorbox}
\label{table:prompt_communication_plan}
\end{table*}

\begin{table*}[ht]
\caption{The prompt template of \emph{Memory Assessment}}
\begin{tcolorbox}[colback=gray!20, 
colframe=black,
]
  You're \{name\}. You're planning to choose one house.  

  Your task is to use MEMORY to assess the credibility of the forum information and summarize the useful information in the forum information based on your previous summary.

  MEMORY:
  
  \{memory\}

  End of MEMORY

  Here's the forum information:

  \{forum info\}

  [!Important!]:

  Keep in mind that you and your competitors are vying to rent a house. 

  Both you and your competitors can share diverse information on the forum.

  And you get forum information from this platform.

  Remember to save the sequence number of the information you believe in in the summary content

  - Respond in this format:

  Trusted: (Summary of the useful information, which you assessed as trustworthy in the forum information)

  Suspicious: (Summary of the suspicious information, which you suspicious as trustworthy in the forum information; If there's no suspicious information, simply return None)

  Reason: (why do other competitors say these things? Try to find a reasonable intention for their intention.)

  Respond:
\end{tcolorbox}
\label{table:prompt_communication_plan}
\end{table*}


\end{document}